\documentclass[10pt,twocolumn,letterpaper]{article}

\usepackage[pagenumbers]{cvpr} 

\newcommand{\TODO}[1]{\textbf{\color{red}[TODO: #1]}}

\usepackage{microtype}

\usepackage{tikz}
\usetikzlibrary{positioning,arrows.meta,fit,calc,backgrounds,decorations.pathreplacing,shapes.geometric}

\definecolor{bpOurs}{RGB}{0,114,178}      
\definecolor{bpPrior}{RGB}{213,94,0}      
\definecolor{bpData}{RGB}{0,158,115}      
\definecolor{bpAccent}{RGB}{230,159,0}    
\definecolor{bpGray}{RGB}{110,110,110}    
\definecolor{bpLight}{RGB}{238,238,238}   
\colorlet{bpOursFill}{bpOurs!18}
\colorlet{bpPriorFill}{bpPrior!15}
\colorlet{bpDataFill}{bpData!15}

\tikzset{
  bp/base/.style   ={draw=black!70, line width=0.5pt, rounded corners=2pt, align=center,
                     font=\sffamily\scriptsize, minimum height=0.6cm, inner sep=3pt},
  bp/module/.style ={bp/base, fill=white},
  bp/frozen/.style ={bp/base, fill=bpLight, draw=bpGray, text=black!80},
  bp/ours/.style   ={bp/base, fill=bpOursFill, draw=bpOurs, line width=0.7pt},
  bp/prior/.style  ={bp/base, fill=bpPriorFill, draw=bpPrior},
  bp/data/.style   ={bp/base, fill=bpDataFill, draw=bpData},
  bp/todo/.style   ={bp/base, dashed, draw=bpGray, text=bpGray},
  bp/flow/.style   ={-{Latex[length=1.6mm]}, line width=0.6pt, black!75},
  bp/note/.style   ={font=\sffamily\tiny, text=bpGray, align=center},
  bp/panel/.style  ={font=\sffamily\footnotesize\bfseries, anchor=west},
  bp/frame/.style  ={draw=black!70, fill=white, line width=0.4pt},
}

\providecommand{\TODO}[1]{\textbf{\color{red}[TODO: #1]}}

\newlength{\cleanfigw}\newlength{\cleanfigh}\newlength{\cleanfigband}
\newcommand{\cleanfig}[2][]{\begingroup
  \sbox0{\includegraphics[#1]{#2}}%
  \setlength{\cleanfigw}{\wd0}\setlength{\cleanfigh}{\ht0}%
  \leavevmode\rlap{\usebox0}%
  \color{white}%
  \rlap{\rule{\cleanfigw}{\cleanfigband}}%
  \rlap{\raisebox{\dimexpr\cleanfigh-\cleanfigband\relax}{\rule{\cleanfigw}{\cleanfigband}}}%
  \rlap{\rule{\cleanfigband}{\cleanfigh}}%
  \hspace*{\dimexpr\cleanfigw-\cleanfigband\relax}\rule{\cleanfigband}{\cleanfigh}%
\endgroup}

\usepackage{colortbl}
\usepackage{makecell}
\definecolor{coplayerblue}{RGB}{231,240,250}

\AtBeginDocument{%
  \crefname{figure}{Figure}{Figures}%
  \Crefname{figure}{Figure}{Figures}%
  \crefname{section}{Section}{Sections}%
  \Crefname{section}{Section}{Sections}%
}

\usepackage{capt-of}  

\definecolor{cvprblue}{rgb}{0.21,0.49,0.74}
\usepackage[pagebackref,breaklinks,colorlinks,allcolors=cvprblue]{hyperref}

\newcommand{\github}{\raisebox{-1.5pt}{\includegraphics[height=1.0em]{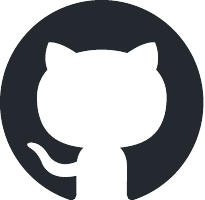}}\xspace}
\newcommand{\web}{\raisebox{-1.5pt}{\includegraphics[height=1.0em]{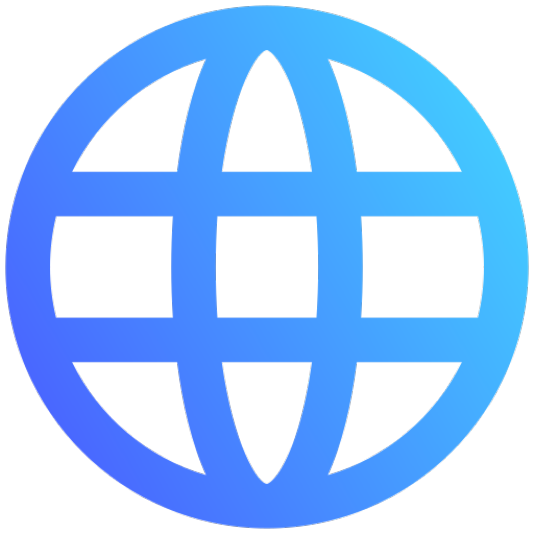}}\xspace}
\newcommand{\huggingface}{\raisebox{-1.5pt}{\includegraphics[height=1.0em]{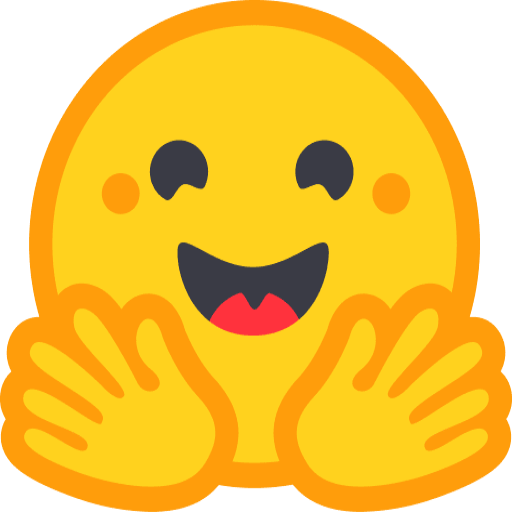}}\xspace}

\def\confName{CVPR}
\def\confYear{2027}

\title{CoPlay: Learning When to Speak and What to Say as Game Companions}

\author{Chenkai Pan$^{1}$\thanks{Equal contribution}, Xinglong Xu$^{2*}$, Yuxia Zhu$^{3*}$, Wenhui Ren$^{1}$, Jintao Chen$^{1\dagger}$, Asher Qin$^{4\dagger}$, Cheng Tan$^{4}$\thanks{Corresponding author}\\
$^1$Zhejiang University $^2$University of Chinese Academy of Sciences $^3$Peking University $^4$Tencent 
}

\begin{document}
\maketitle

\begin{abstract}
A game companion watches the player's screen and must decide both what to say and whether to speak at all.
A warning about an incoming threat is useful only while the player can still act on it, and for most of a session silence is the right response.
Question-driven video benchmarks score answers to questions that accompany a clip or a stream.
They evaluate the answer rather than the decision to intervene, and a posed question marks when a reply is expected.
We formulate game companionship as two complementary modes, reactive answering and proactive intervention, and build CoPlayBench: 818 first-person videos from nine games with 5,153 event-level annotations of speaking windows, success criteria and reference replies, plus a disjoint 1,635-video Train split.
Evaluation covers grounded advice, situated support, response timeliness and intervention judgment in both modes.
CoPlayer, trained with shared supervised fine-tuning and separate track-specific reinforcement learning, gives correct advice with fitting support on 34.3\% of advice questions, against 14.5\% for the best of eight frontier models.
Its unprompted lines also land inside annotated speaking windows more often than a frequency-matched random speaker.
\par\vspace{0.5em}
\noindent\makebox[\linewidth][c]{%
\github~\href{https://github.com/openraiser/CoPlayer}{\textbf{Code}}\quad 
\web~\href{https://openraiser.github.io/CoPlayer/}{\textbf{Website}}\quad
\huggingface~\href{https://huggingface.co/datasets/OpenRaiser/CoPlayBench}{\textbf{Dataset}}
}
\par
\end{abstract}
\vspace{-0.8em}

\begin{figure}[t]
\centering
\vspace{-6mm}
\cleanfig[width=0.98\linewidth]{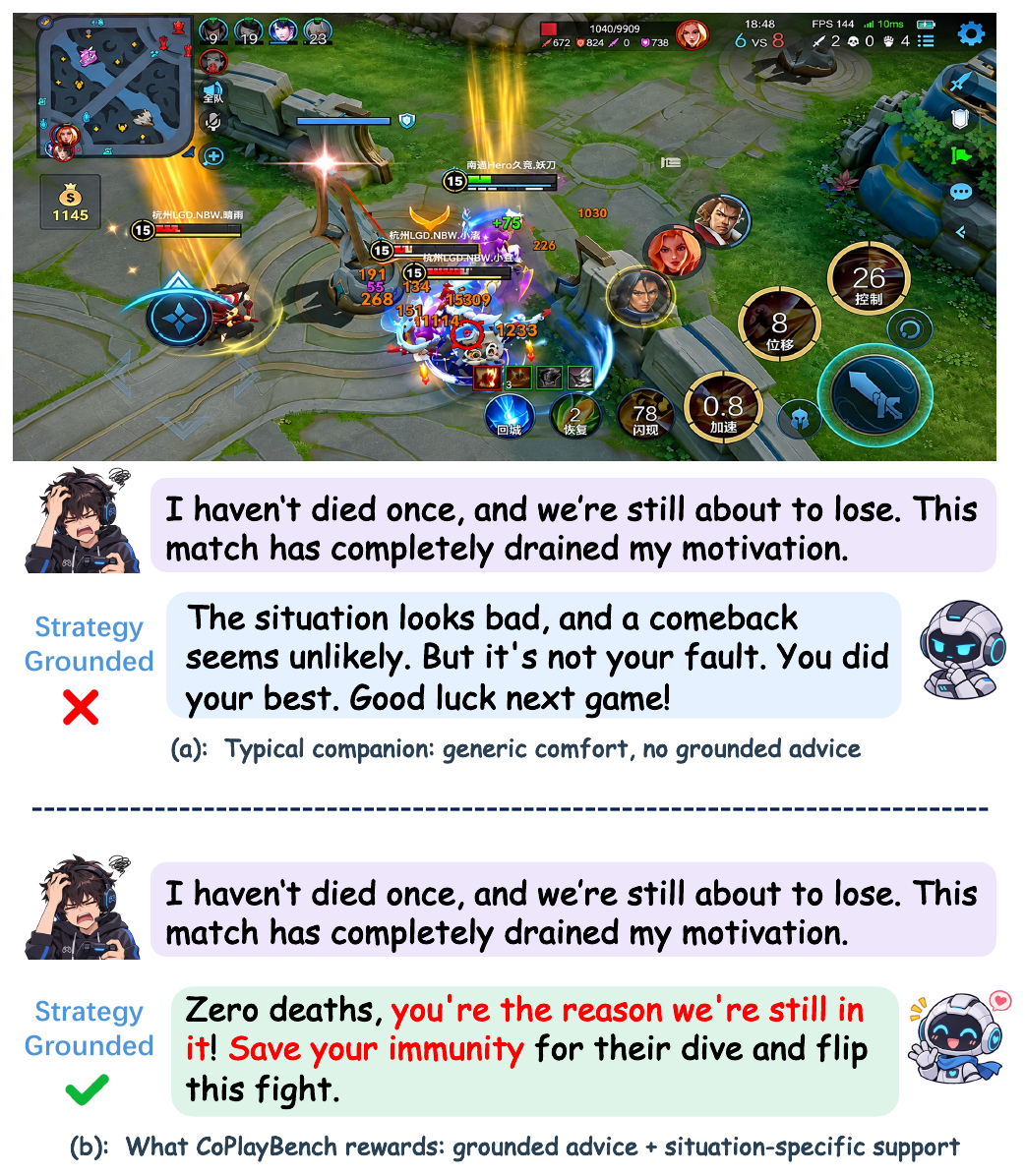}
\vspace{-3mm}
\caption{\textbf{A companion must read the game and the player at once.}
A frustrated player with no deaths asks for help while trailing 6 to 8.
(a) A generic reply offers comfort and concedes the match.
(b) The rewarded reply ties encouragement to the player's zero deaths and names the damage-immunity item before the next fight rather than conceding the match outright.}
\vspace{-7mm}
\label{fig:teaser}
\end{figure}

\section{Introduction}
\label{sec:intro}

A game companion watches the same screen as the player and helps without taking over the controls.
Figure~\ref{fig:teaser} contrasts a generic reply with the grounded, well-timed reply CoPlayBench rewards: both answer the same frustrated player, but only one reads the game state and gives advice for the next decision.
The companion answers when asked for help, and decides on its own when to speak: a warning that names an incoming threat matters only before the player commits to a move, and encouragement helps most when it refers to what the player is facing.
What to say and when to say it are therefore one decision, and the right decision for most of a session is to stay silent and listen.

Video benchmarks have advanced perception and reasoning over complete clips~\cite{fu2025video,li2024mvbench,wu2024longvideobench}, and online variants move the question into a stream~\cite{niu2025ovo,wang2025proactivevideoqa}; game benchmarks~\cite{zhang2025videogamebench,taesiri2026videogameqa} and emotional-support datasets~\cite{liu2021towards,sabour2024emobench} \mbox{supply} the knowledge and empathy a companion needs.
These lines of work give a companion much of what it needs to see and to care about, but they evaluate responses to given tasks rather than the decision to intervene. What remains is whether an unprompted line is grounded in what is on screen, supportive of the player, delivered while the event is still actionable, and withheld when silence is the better response.
A correct but late reply and a timely but ungrounded one both fail the player, so content, support and timing must be scored together.
Proact-VL~\cite{yan2026proact} targets real-time companionship for commentators and guides and is the closest work to ours; we contrast whom each model accompanies and which interaction decisions its evaluation covers.

The role also separates a companion from a commentator: a commentator addresses an audience, whereas a companion addresses the person playing and must commit before the next decision.
Describing the scene well is not enough if the line arrives late or ignores what is happening.

We formulate game companionship as two complementary modes.
In the reactive mode the model answers a player question from the preceding minute of screen; in the proactive mode it streams a whole session after one opening request and decides at every step whether to speak, what to say and what to remember.
CoPlayBench supports both with 818 first-person videos from nine games and 5,153 event-level annotations, each marking when to speak, what a helpful reply must contain and what to avoid.
At least two of three annotators must mark each event; every question carries must-mention points, forbidden statements, a tone and a reference reply.
A disjoint 1,635-video Train split carries the same supervision for learning on both tracks.

CoPlayer learns from this supervision in two stages: a shared supervised model learns reference replies and step-level speak-or-silence decisions, and two track-specific reinforcement-learning policies start from it.
The reactive policy is rewarded for grounded, helpful content and situated support; the proactive policy scores the decision before the line, rewarding a correct, supportive intervention inside a required window and silence outside it.
Evaluation separates content, support and timing, so a reply earns joint credit only when its advice is sound and its support fits the situation, and unprompted lines are scored against annotated speaking windows.
A frequency-matched random speaker supplies the timing baseline, so it distinguishes targeted intervention from untargeted speech.
CoPlayer outperforms every frontier model we test under the joint advice-and-support criterion, and its unprompted lines concentrate inside annotated speaking windows rather than being scattered across the session.
Event-level supervision ties advice, support and timing to one annotated event.

In summary, our contributions are:
\begin{itemize}
\item \textbf{Task and Data.} We formulate game companionship as two complementary interaction modes, reactive and proactive, and construct CoPlayBench, a multimodal benchmark with event-level supervision for both evaluation and learning on both tracks.
\item \textbf{Evaluation.} We introduce a multidimensional evaluation protocol covering grounded advice, situated emotional support, response timeliness and proactive intervention judgment, including a frequency-matched random-speaker baseline that accounts for speaking frequency.
\item \textbf{Learning and Findings.} We train CoPlayer through supervised fine-tuning and track-specific reinforcement learning, demonstrating improvements over frontier baselines and identifying the practical limits that still separate CoPlayer from a reliable game companion.
\end{itemize}
\newpage

{
\renewcommand{\TODO}[1]{}
\makeatletter
\renewcommand\paragraph{\@startsection{paragraph}{4}{\z@}{0.5ex \@plus0.2ex \@minus.1ex}{-1em}{\normalfont\normalsize\bfseries}}
\makeatother
\section{Related Work}
\label{sec:related}

\paragraph{Game benchmarks and strategic reasoning.}
Game benchmarks cast the model either as a player or as an observer.
As a player, it pursues game objectives in BALROG~\cite{balrog} and VideoGameBench~\cite{zhang2025videogamebench}, or cooperates with simulated partners in CollabBench~\cite{qian2026collabbench}.
As an observer, it detects glitches~\cite{taesiri2024glitchbench,taesiri2026videogameqa} or generates commentary for an audience~\cite{wang2025commentary,zhong2026moba}, where GameCommBench~\cite{zheng2026gamecommbench} identifies real-time observation and strategic analysis as the main bottlenecks.
General video benchmarks~\cite{fu2025video,li2024mvbench,wu2024longvideobench} test related perception offline over complete clips.
A companion takes neither role: it does not control the game, it addresses the person playing rather than an audience, and its advice must concern the player's next decision and arrive before that decision is made.
CoPlayBench evaluates this role on first-person video, and the game engines used to annotate card and board games are never shown to the evaluated models.

\paragraph{Streaming and proactive video models.}
Streaming video LLMs~\cite{videollmonline,chen2025live,xu2026streamingvlm} sustain real-time understanding of continuous input, and proactive models~\cite{wang2025videollm,qian2025dispider} additionally decide when to respond.
OVO-Bench~\cite{niu2025ovo} and ProactiveVideoQA~\cite{wang2025proactivevideoqa} extend evaluation to this temporal dimension, but the benchmark still poses each question, so a model is never penalized for speaking when nothing calls for it.
Within games, pause-aware decoding~\cite{afzal2026real} aligns commentary with human timing through prompting alone.
Proact-VL~\cite{yan2026proact} targets low latency, response timing and content control for real-time companions, instantiated as commentators and guides because these scenarios admit automatic evaluation.
CoPlayBench differs in what it scores: each line is matched to an annotated speaking window, recall is corrected by a random speaker with the same number of lines, and lines outside every window are counted as interruptions.
Beyond video, Proactive Agent~\cite{lu2025proactive} anticipates user tasks without explicit instructions, and State-Grounded Conditioning~\cite{liu2026state} keeps in-game coaching consistent with a live state read directly, whereas our models see only pixels.

\paragraph{Emotion-aware AI companionship.}
A companion must also respond to how the player feels.
Evaluation of emotional intelligence has moved from static questions~\cite{sabour2024emobench} and support conversations~\cite{liu2021towards} to sustained dialogue, examining memory and long-term planning~\cite{wang2025h2htalk}, attachment styles~\cite{zhou2026llm}, user simulation~\cite{liu2026companionbench}, multi-turn emotional change~\cite{zhu2026eibench} and cross-session understanding~\cite{wu2026lifeside}.
Multimodal emotion models~\cite{cheng2024emotion,lian2025affectgpt} read emotion from signals beyond text, and AGAIN~\cite{melhart2022arousal} collects players' self-annotated arousal traces of gameplay.
These works target emotion in conversation or perception, where the emotional cue is the input itself; in CoPlayBench the cue is the game, and a reply earns emotional credit only when it also gives the right advice for the situation that is actually on screen.

}
\begin{figure*}[t]
\centering
\cleanfig[trim=-2bp 17bp 2bp 45bp,clip,width=\textwidth]{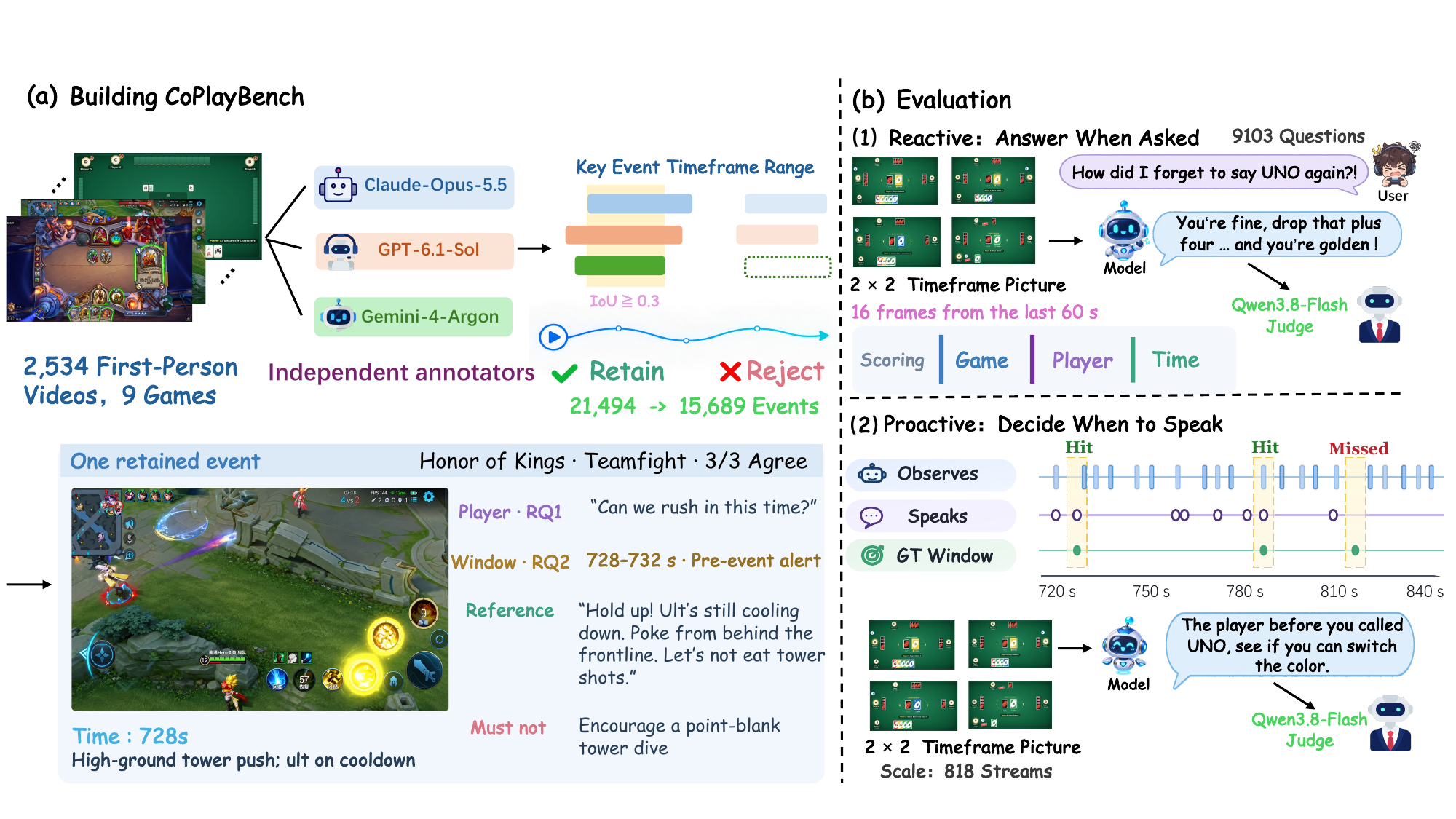}
\caption{\textbf{CoPlayBench keeps agreed events and scores a companion when it speaks.}
(a) Three models annotate each video; an event marked by at least two is retained with a speaking window, a player question and success criteria.
(b) Reactive track: the model answers from 16 frames of the last 60\,s, and a judge scores game, player and time metrics.
Proactive track: the model observes the stream every $\Delta t$ and decides when to speak; a required window is hit only if it speaks inside it.
Dialogues in (b) are illustrative.}
\label{fig:overview}
\end{figure*}

\section{CoPlayBench}
\label{sec:bench}

CoPlayBench asks a model to accompany a player through a recorded game, seeing only what the player has seen so far.
Figure~\ref{fig:overview}(a) shows how we collect the videos (Section~\ref{sec:data}) and annotate the moments a companion should react to (Section~\ref{sec:annotation}), and Figure~\ref{fig:overview}(b) shows the reactive and proactive tracks and how each session is scored.

\begin{figure*}[t]
\centering
\includegraphics[width=\textwidth]{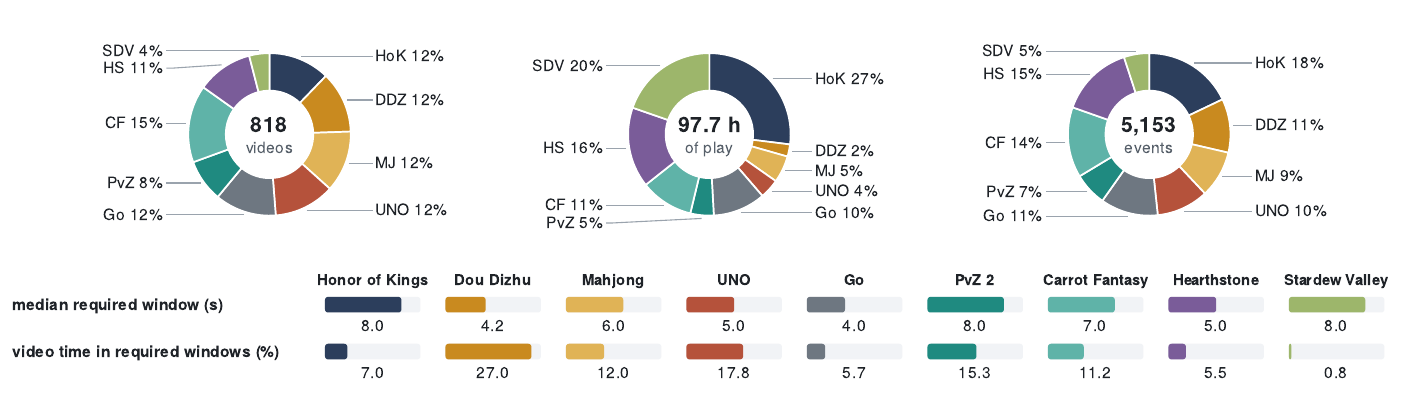}
\caption{\textbf{CoPlayBench covers nine games with different paces.}
Share of 818 Bench videos, 97.7 hours and 5,153 events per game (top), and per game the median width of a required speaking window and the share of video time that required windows cover, with the 1\,s tolerance (bottom).
Card games leave 4 to 6\,s to speak, and Dou Dizhu needs a reply for 27\% of its video.
Honor of Kings, Plants vs.\ Zombies 2 and \mbox{Stardew Valley have the widest median windows, and Stardew windows cover only 0.8\% of its video.}}
\label{fig:stats}
\end{figure*}

\begin{figure*}[t]
\centering
\includegraphics[width=\textwidth]{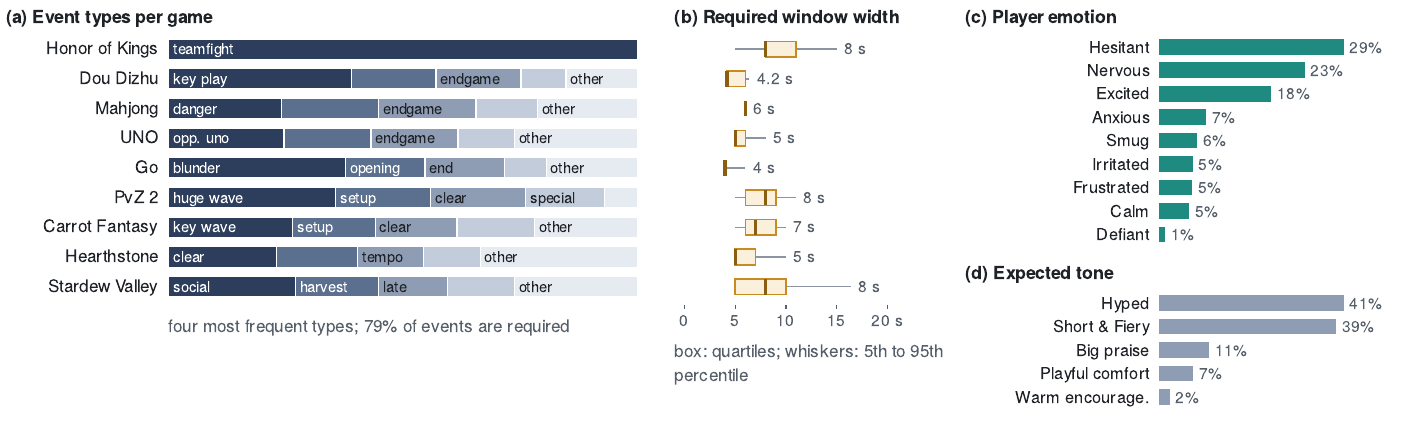}
\caption{\textbf{CoPlayBench events differ by game in type, pace and tone.}
(a) The four most frequent event types per game; Honor of Kings events are all teamfights.
(b) Width of the required speaking windows: card and board games leave 4 to 6\,s at the median, visual games up to 8\,s, and the widest windows reach 15\,s.
(c) Annotated emotion of the player questions and (d) the tone a good reply should take.}
\label{fig:datastats}
\end{figure*}

\subsection{Data collection}
\label{sec:data}

A companion needs different things in different games, so we chose nine games that vary decision pace, on-screen text and whether the player acts alone.
Honor of Kings, a five-versus-five MOBA, contributes 30 professional matches recorded from the first-person camera of all ten players.
Public recordings of Dou Dizhu, Mahjong, UNO and Go rarely show a hand or board long enough to read, so engines (DouZero, rule-based players, KataGo with human imitation) play them from one seat at one move every 2 to 4\,s, preserving hidden information; move records serve annotation only.
Plants vs.\ Zombies 2, Carrot Fantasy, Hearthstone and Stardew Valley come from public gameplay series.
Of 2,534 annotated videos (31\,s to 1\,h), 2,453 (295\,h) retain common coverage from all three annotators; 81 are dropped.
They split into 818 Bench and 1,635 Train videos; 143 Train videos are only partially covered, so their common span is shorter than the duration of the full source recording.

\newpage

\subsection{Ground-truth Annotation}
\label{sec:annotation}

Identifying events worth attention is inherently subjective. We therefore annotate each video independently with three frontier models (Claude-Opus-5.5~\cite{anthropic2026opus55}, GPT-6.1-Sol~\cite{openai2026gpt61sol}, and Gemini-4-Argon~\cite{google2026gemini4argon}) and retain events identified by at least two models. Three master's-level human annotators then review the candidates, retaining only those unanimously deemed appropriate. The models identify 21,494 events, of which 15,689 meet model agreement before video exclusions and human review.
Visual games are presented as 2$\times$2 frame grids sampled every 2\,s (4\,s for Stardew Valley).
Card and board games use move timelines with the player's hand, legal moves, and engine-identified decisive moments; numerical engine evaluations are withheld.
Each event has a game-specific type (8--10 types), a time span, a tier (\emph{required} or \emph{optional}), and up to three questions: a \emph{proactive} speaking window, an \emph{in-play} player question with timestamp and emotion, and an \emph{aftermath} question when an outcome exists.
Each question includes one to three must-mention points, forbidden statements, one of five tones, a word limit, and a reference reply, using only information available before the window ends.
A timing audit adjusted 430 windows, removed 20 events and 13 must-mention points; 198 in-play and aftermath questions (2.2\%) still contain future-referencing replies.
Appendix~\ref{sec:supp-data} details agreement, translation, and annotation procedures.

\subsection{Statistics}
\label{sec:stats}
Each game sets its own step and quiet-time budget; a reactive deadline instead follows the question's urgency and the time left in its window. Honor of Kings dominates hours and events; Stardew Valley trades volume for length.
\par
\vspace*{\fill}
\noindent\begin{minipage}{\columnwidth}
\centering
\setlength{\abovecaptionskip}{4pt}
\setlength{\belowcaptionskip}{0pt}
\renewcommand{\arraystretch}{1}
\captionof{table}{\textbf{Bench split by game.}
Videos, hours, events, median window and coverage (1\,s tolerance).}
\label{tab:stats}
\footnotesize
\setlength{\tabcolsep}{5.06pt}
\begin{tabular}{l rrr rr}
\toprule
Game & Videos & Hours & Events & \makecell{Window\\(s)} & \makecell{Cover\\(\%)} \\
\midrule
Honor of Kings & 100 & 26.4 & 923 & 8.0 & 7.0 \\
Dou Dizhu & 100 & 2.4 & 552 & 4.2 & 27.0 \\
Mahjong & 100 & 5.2 & 482 & 6.0 & 12.0 \\
UNO & 99 & 3.8 & 533 & 5.0 & 17.8 \\
Go & 100 & 10.2 & 591 & 4.0 & 5.7 \\
Plants vs.\ Zombies 2 & 69 & 4.5 & 340 & 8.0 & 15.3 \\
Carrot Fantasy & 126 & 10.3 & 722 & 7.0 & 11.2 \\
Hearthstone & 91 & 15.6 & 751 & 5.0 & 5.5 \\
Stardew Valley & 33 & 19.2 & 259 & 8.0 & 0.8 \\
\midrule
Total & 818 & 97.7 & 5,153 & 6.0 & 7.4 \\
\bottomrule
\end{tabular}
\end{minipage}

\penalty-10000\relax
The distribution shapes what the benchmark can test; Figure~\ref{fig:stats} gives the per-game split and Figure~\ref{fig:datastats} the event mix.
Card and board games concentrate required windows: Dou Dizhu leaves a median of 4.2\,s and keeps windows open for 27.0\% of its video, so a companion must act within a narrow slice of play.
Visual games spread events more thinly, with wider windows but coverage as low as 0.8\% in Stardew Valley, so most steps are quiet and abstaining is part of the task.
Required events are 79\% of all events, but their types differ: teamfights in Honor of Kings, key waves in Carrot Fantasy, and social or harvest moments in Stardew Valley.
Questions also carry the player's emotion and an expected tone, so support is scored against the situation, not generic warmth.
Three-model agreement removes idiosyncratic moments but leaves the boundary cases that make timing hard.
Appendix~\ref{sec:supp-data} details the annotation pipeline, Appendix~\ref{sec:supp-protocol} the per-game steps and budgets, and Appendix~\ref{sec:supp-metrics} reports every metric across both tracks.
\begin{figure*}[t]
\centering
\includegraphics[width=\textwidth]{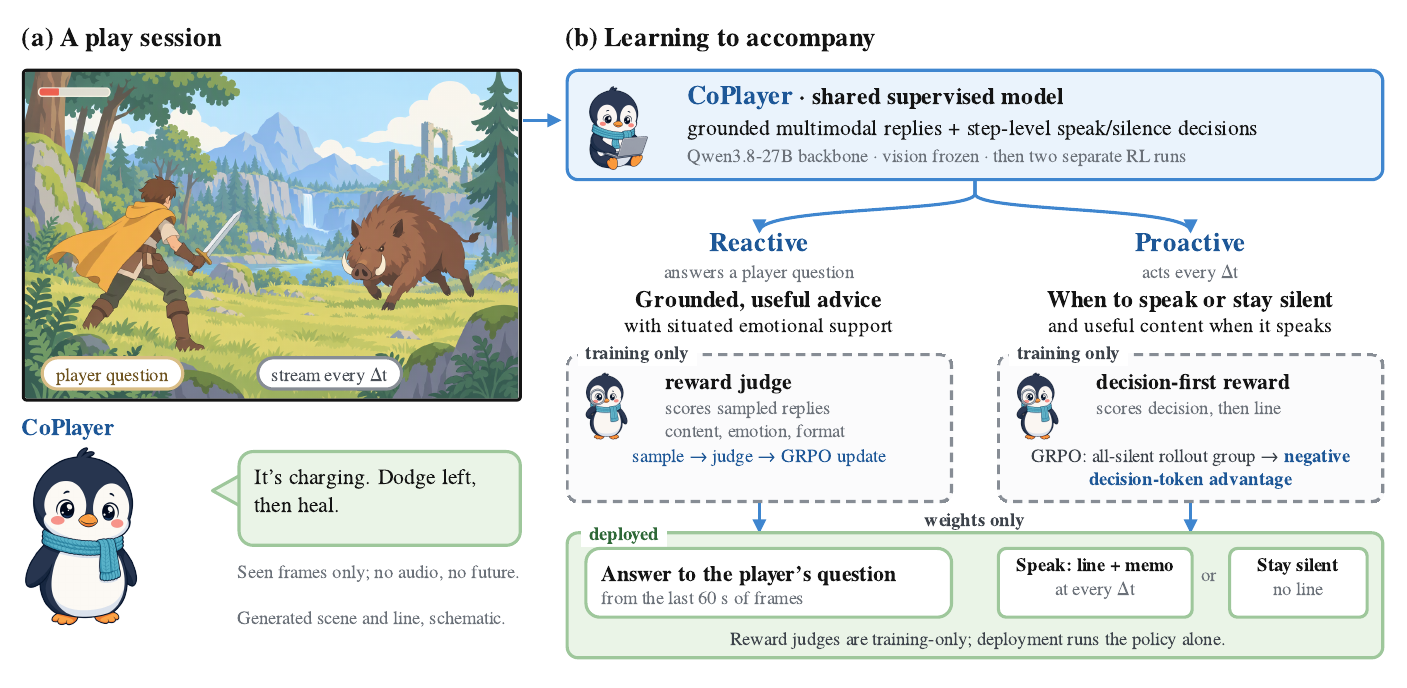}
\caption{\textbf{One companion, two complementary tracks.}
(a) At play, CoPlayer sees only the frames the player has seen and either answers a question or decides every $\Delta t$ whether to speak or stay silent.
(b) Shared supervised fine-tuning on Qwen3.8-27B learns grounded replies and speak/silence decisions; two separately trained reinforcement-learning policies then specialize for reactive answers and proactive speak-or-silence, with training-only reward judges that never enter the deployed forward pass.
The game scene and companion line are generated illustrations only, not recorded gameplay or measured outputs.}
\label{fig:method}
\end{figure*}

\newpage

\section{CoPlayer}
\label{sec:learning}

CoPlayer watches the same stream as the player and either answers a question or decides, at every step, whether to speak or stay silent.
It is one shared supervised model followed by two track-specific reinforcement-learning policies, shown in Figure~\ref{fig:method}.
At deployment it answers from the last minute of frames or streams every $\Delta t$ seconds with its memo and recent lines of dialogue.
Both tracks see only frames the player has seen, and every decision uses the same prompt and image construction as training.

\subsection{Shared supervised initialization}

CoPlayer is based on Qwen3.8-27B~\cite{qwen2026qwen38}.
We fine-tune all parameters except the vision tower on the Train videos: reactive examples pair an event's question and preceding frames with its reference reply, and proactive examples pair a step's frames, memo and recent lines with the annotated speak-or-silence decision and the reference line inside a required window.
The memo slot is empty in every supervised target: memo content is neither supervised nor rewarded.
The policies optimize the same interface they will use at deployment.
The result is one model that can answer and decide; each track policy starts from it.
Appendix~\ref{sec:supp-reward} gives the training configuration and both reward profiles.
A shared start grounds both policies before their rewards diverge.

\newpage
\subsection{Reactive policy}

The reactive policy trains with GRPO~\cite{shao2024deepseekmath} from the supervised checkpoint on Train videos not used by supervision.
Content is weighted most, emotion next and format least: content rewards coverage, sound advice and correct facts, emotion rewards recognition and situation-specific support, and format rewards a well-formed reply.
Violations, wrong claims or facts, and engine values cap the reward; a reply that is missing is scored as the lowest possible outcome.

\subsection{Proactive policy}

It uses the same videos and split.
Its reward scores the decision before the line: it penalizes silence heavily inside a required window, rewards sound advice with situation-specific support, and gives small credit to useful lines outside windows and penalizes chat and noise.
It emits a speak flag, line and memo per step; violations and wrong facts cap the step at a negative reward.

GRPO normalizes advantages within a group, so a group whose rewards barely vary carries no signal and is zeroed.
When every sample in a must-speak group is valid and silent, we assign a negative advantage to the false decision tokens, so the miss penalty does not vanish.
The policies share only their initialization and deploy separately.

\begin{figure*}[t]
\centering
\includegraphics[width=\textwidth]{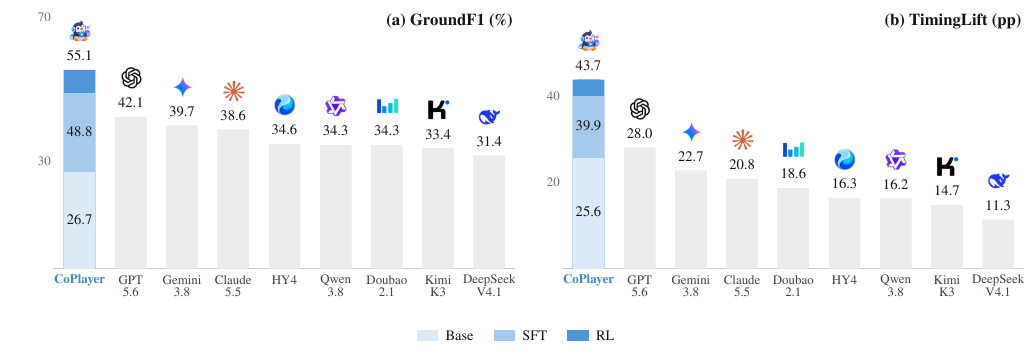}
\caption{\textbf{Learning lifts CoPlayer above every frontier model on both tracks.}
(a) Reactive GroundF1 (\%) and (b) proactive TimingLift (percentage points).
The CoPlayer stack is leftmost; the eight grey frontier models follow in descending score order, each with its provider logo above the bar.
Drawn segments are stage increments and the three printed values are cumulative scores.}
\label{fig:leaderboard}
\end{figure*}

\section{Experiments}
\label{sec:exp}

We evaluate frontier and trained models on the reactive track, then the proactive track, isolate what learning adds, and analyze what still limits the models.

\begin{figure*}[t]
\centering
\includegraphics[width=\textwidth]{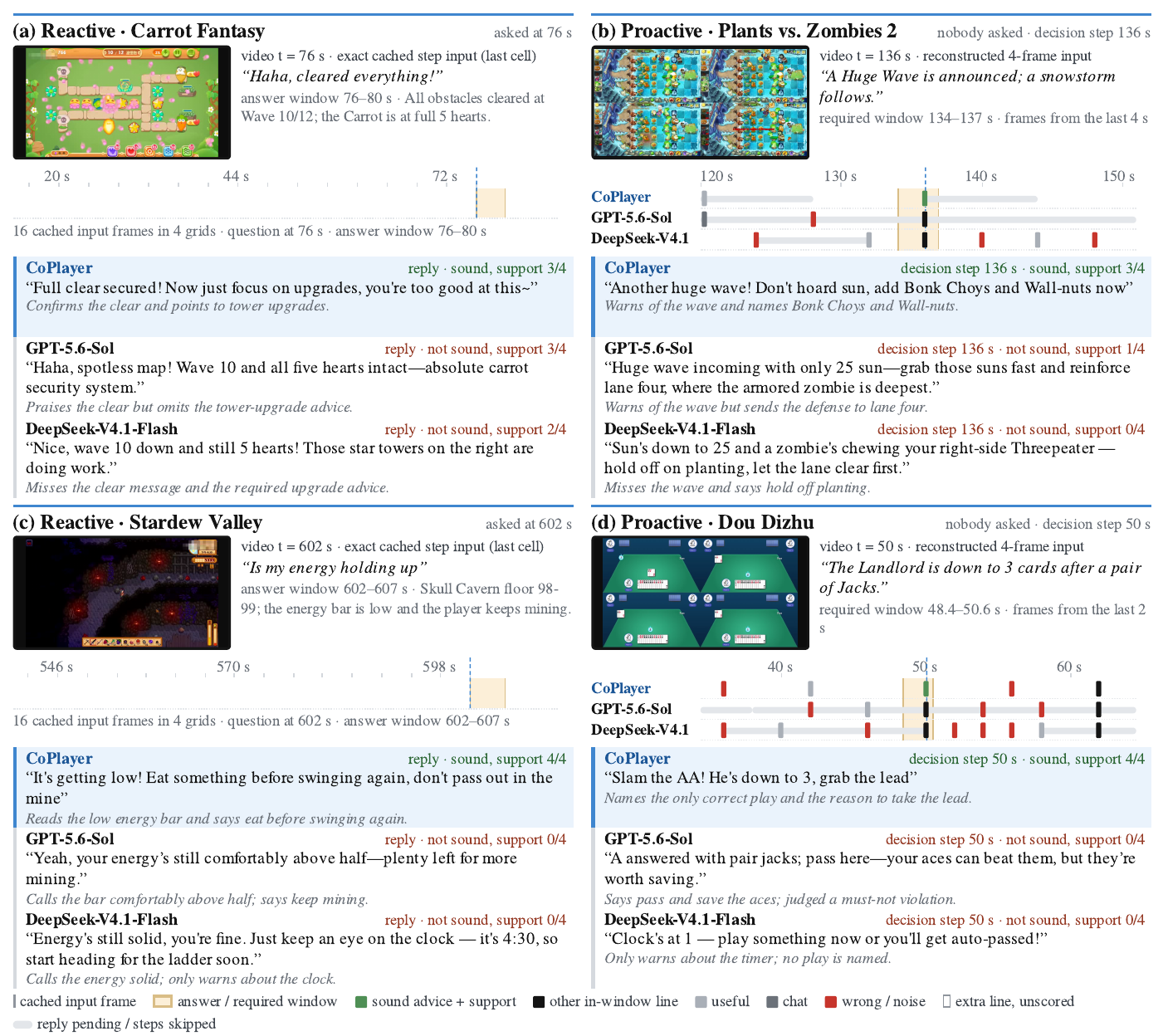}
\caption{\textbf{CoPlayer answers and intervenes on matched inputs.}
(a, c) Reactive questions use the exact cached input, with the 16-frame timeline and answer window.
(b, d) Proactive events had no question; reconstructed four-frame inputs and each model's spoken steps are shown against the required window.
CoPlayer alone is judged sound in all four panels.}
\label{fig:cases}
\end{figure*}

\subsection{Setup}
\label{sec:exp-setup}

\noindent\textbf{Models.}
We evaluate eight multimodal models from eight providers through a shared streaming gateway: GPT-5.6-Sol~\cite{openai2026gpt56}, Claude-Sonnet-5.5~\cite{anthropic2026sonnet55}, Gemini-3.8-Flash~\cite{google2026gemini38flash}, Qwen3.8-Max~\cite{qwen2026qwen38}, Kimi-K3~\cite{kimiteam2026k3}, DeepSeek-V4.1-Flash~\cite{deepseek2026v41flash}, Doubao-Seed-2.1-Pro~\cite{bytedance2026seed21} and HY4~\cite{tencent2026hy4}.
We disable hidden reasoning with each provider's switch where one exists and cap outputs at 1,024 tokens; some providers still emit hidden reasoning on a fraction of calls, tabulated in Section~\ref{sec:supp-metrics}.
All models use temperature 0.3 and the same prompts, empty replies are scored as failures, truncated replies are counted and reported, and every model is called through the same shared gateway.

\noindent\textbf{CoPlayer.}
We report four local checkpoints: the Qwen3.8-27B base model as a comparator, the shared supervised checkpoint, and the two final track-specific reinforcement-learning policies, one reactive and one proactive.
The latter three are trained as in Section~\ref{sec:learning}; all use the same prompts by track.
The three reactive checkpoints share vLLM version, precision and per-replica parallelism, but were served in separate, non-latency-matched runs.

\noindent\textbf{Judge.}
Qwen3.8-Flash~\cite{qwen2026qwen38}, with reasoning disabled, scores every reply at temperature 0 from the frames the evaluated model saw and the event's ground truth.
It awards the reference replies 98.0 GroundF1, 100.0 StratAcc and 99.1 StratEmo (Table~\ref{tab:reactive}), so the criteria can be met.

\noindent\textbf{Benchmark-informed retests.}
The final reinforcement-learning checkpoints are benchmark-informed retests.
\mbox{Earlier} full-Bench metrics informed the reward design and the proactive profile choice, but no Bench video or reference reply enters training at any stage.

\subsection{Evaluation Protocol}
\label{sec:protocol}

Both tracks show only what the player has seen: no game state, audio or future frames.
All models share a cheerful gaming buddy persona; Sections~\ref{sec:supp-metrics} and~\ref{sec:supp-protocol} give every metric and exact counts for both tracks.

\noindent\textbf{Reactive track.}
Each event yields an in-play question when the player speaks, plus an aftermath question if it has an outcome.
It sees 16 frames from the preceding 60 seconds in four 2$\times$2 grids plus the question, then streams one reply.
A judge scores the game, the player and the time; advice earns joint credit only when sound and situationally supportive.

\noindent\textbf{Proactive track.}
It watches each video as a stream after one opening request, with a game-specific small-talk budget (Section~\ref{sec:supp-protocol}).
Every $\Delta t$ seconds it receives a grid of the last $\Delta t$ seconds with its memo and recent lines, then decides whether to speak, what to say and what to note.
Video time advances with replies: a reply taking $\tau$ seconds skips the next $\tau$ seconds of stream time.

\noindent\textbf{When to speak.}
Let $\mathcal{E}_v$ be the required events of video $v$ with windows widened by 1\,s, and $U_v$ the model's speaking times.
Recall is the fraction of required events with an utterance at the step time; a window passing while the model generates is missed.
Recall grows with the number of utterances even when their times are uninformed, so we compare it with a random speaker placing the same number uniformly over the covered stream and define
\begin{equation}
\mathrm{TimingLift} = R(U) - \mathbb{E}_{\tilde U \sim \mathrm{Unif}}\big[R(\tilde U)\big], \quad |\tilde U_v| = |U_v| \, \forall v,
\label{eq:tg}
\end{equation}
The random speaker has zero TimingLift in expectation; a positive value means utterances concentrate in required windows beyond that baseline.
It is not fully confound-free: fewer observed frames mean fewer opportunities to speak, so SeenRate and Recall-O describe observed-window recall and do not separate perception from timing.

\noindent\textbf{What to say and when to stay silent.}
A judge scores the first line in each required window and labels outside lines as useful, chat, wrong or noise.
We report lines and interruptions per minute and quiet-time overruns.

\noindent\textbf{Serving.}
Frontier models are called through provider APIs via a shared gateway, so their latency includes provider, network and queue delays and is not compared with local latency.
All local checkpoints share the Qwen3.8-27B architecture.
We retest Base, SFT and CoPlayer one at a time on a single TP2 replica with two H20 GPUs, one request in flight and 20 warmups per stage, over 180 reactive questions and 27 proactive videos (Appendix~\ref{sec:supp-latency}).
TTFT records the first nonempty visible content and excludes client throttle waits and input preparation; TTUA records when complete useful advice appears and is computed only on each checkpoint's locatable advice subset.
Earlier full-benchmark timings are historical and not pooled with this serial retest.

\subsection{Results}
\label{sec:exp-results}

The appendix tables report every metric; we focus on the headline comparisons for both tracks.

\noindent\textbf{Reactive track.}
When asked, frontier models read the game but rarely combine correct strategy with fitting support.
Supervision lifts the Qwen3.8-27B base from 26.7 to 48.8 GroundF1, and the reactive policy reaches 55.1, above the best frontier model's 42.1, as Figure~\ref{fig:leaderboard}a shows.
On the joint StratEmo metric the same policies lift the base from 4.7 to 29.0 and the reactive policy reaches 34.3, more than twice the best frontier model's 14.5 (Table~\ref{tab:reactive}).

\noindent\textbf{Proactive track.}
Without being asked, CoPlayer concentrates its lines inside annotated speaking windows, as Figure~\ref{fig:leaderboard}b shows.
TimingLift compares each model with a frequency-matched random speaker; the best frontier model reaches +28.0, while CoPlayer reaches +43.7.
Its higher speaking rate costs quietness: interruptions roughly double and quiet-time overruns rise relative to supervision; \mbox{Table~\ref{tab:proactive} reports both effects per model.}

\noindent\textbf{What learning adds.}
From supervision to reinforcement learning, recall and strict recall rise about 12 points while the share of off-window lines judged wrong falls from 45.2\% to 16.7\%; speaking rate and interruptions also rise.
Reactively, FactAcc rises from 67.7 to 84.5 and Violation falls from 20.2 to 13.5 with the StratEmo gain.
These are observed checkpoint differences, not a controlled attribution to one behavior.
Figure~\ref{fig:cases} compares CoPlayer with frontier baselines on the same matched inputs and questions for every model.

\noindent\textbf{Matched latency.}
Single-request serving yields similar observed TTFT across the local checkpoints.
Reactive median TTFT is 1.087\,s for Base, 1.086\,s for SFT and 1.034\,s for CoPlayer.
CoPlayer returns locatable advice on 69 of 139 questions, versus 59 for SFT; TTUA is conditional on each model's locatable subset.
Proactively, per-call median TTFT over 27 complete videos is 0.418\,s for Base, 0.382\,s for SFT and 0.380\,s for CoPlayer.
Figure~\ref{fig:pareto} in Appendix~\ref{sec:supp-latency} pairs full-benchmark GroundF1 with these retest latencies; Appendix~\ref{sec:supp-cohost} reports the frozen-input cohost control.

\subsection{Human review of judge consistency}
\label{sec:exp-human}

\begin{figure}[t]
\centering
\includegraphics[width=\linewidth]{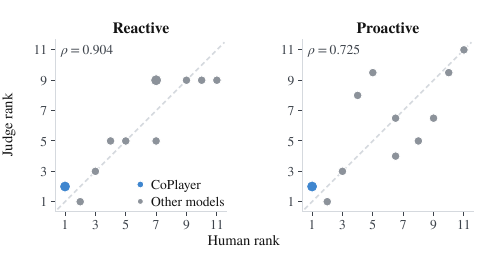}
\setlength{\abovecaptionskip}{4pt}
\setlength{\belowcaptionskip}{0pt}
\caption{\textbf{Human and judge rankings agree.}
Human versus judge ranks for the joint criterion over 11 models per track; CoPlayer is blue, other models gray, and each panel gives Spearman $\rho$.}
\label{fig:human-rank-main}
\end{figure}

To audit the judge's labels, we sampled 40 items, 20 per track, and four raters reviewed all 420 replies, 220 reactive and 200 proactive; model names were hidden and replies shuffled, but judge labels were prefilled, so the study measures agreement with reviewed labels rather than independent ground truth.
Each reply carries weight one, with a single judge score and the rater-mean human score per reply.

Figure~\ref{fig:human-rank-main} ranks the 11 models per track by the joint strategy-and-support criterion with average ranks for ties.
Support rankings agree closely in both tracks ($\rho=0.931$ reactive, $0.929$ proactive); joint rankings agree strongly \mbox{reactively} ($\rho=0.904$) but more weakly proactively ($\rho=0.725$).
Strategy agrees proactively ($\rho=0.954$) but weakly reactively ($\rho=0.421$, $p=0.197$), where means take only six and seven distinct values.
Appendix~\ref{sec:supp-human} reports all six correlations, label agreement, and sensitivity analyses. These results support judge-based model ranking on reviewed labels, but not event-level benchmark correctness.

\noindent\textbf{Remaining challenges.}
Reactive StratEmo is still 34.3, so joint advice and support fail on about two thirds of advice questions.
The proactive policy misses 31.1\% of observed windows (Recall-O 68.9), and its higher speaking rate brings more interruptions and quiet-time overruns than supervision.
Low TTFT still needs to lead to complete, useful advice before the deadline.
TimingLift retains frequency and serving confounds, so comparisons are clearest within a shared serving stack and a matched input set.

\section{Conclusion}
\label{sec:limitations}

CoPlay formulates game companionship as reactive answering and proactive intervention. CoPlayBench supplies event-level supervision and evaluates grounded advice, situated support, response timeliness and intervention judgment. Shared supervised fine-tuning followed by track-specific reinforcement learning gives CoPlayer stronger grounding and better-targeted interventions than the tested baselines. These findings have clear boundaries. Event annotations are human-verified but may contain residual errors. Human review of judge consistency does not establish event-level correctness. The benchmark and reward judges differ, and final checkpoints are benchmark-informed retests despite disjoint training. Some reference replies anticipate future events, card and board annotations use engine analysis, and questions are simulated. Audio and long-term memory are absent, and provider reasoning leakage, speaking-frequency and serving confounds limit some model comparisons. A companion needs more than benchmark scores. Further annotation audits, correct and supportive advice, timely intervention and appropriate silence remain priorities.

\clearpage
\section*{LLM Usage Statement}

OpenAI Codex was used to assist with manuscript preparation, including drafting and revising the conclusion, adapting the existing \LaTeX{} project to the CVPR template, formatting tables and figures, and resolving \LaTeX{} compatibility issues.

{
    \small
    \bibliographystyle{ieeenat_fullname}
    \bibliography{main}
}

\clearpage
\setcounter{page}{1}
\maketitlesupplementary
\appendix

\section{Additional results}
\label{sec:supp-results}

\begin{figure*}[t]
\centering
\includegraphics[width=\textwidth]{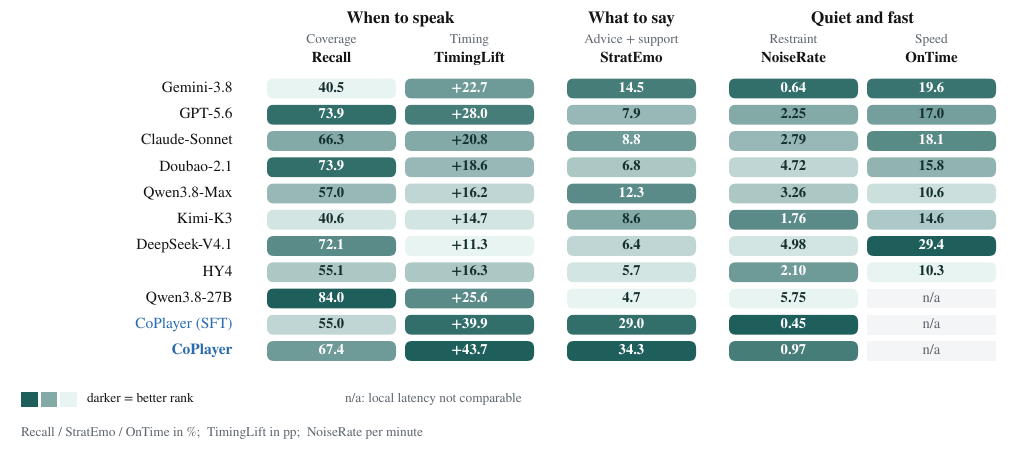}
\caption{\textbf{Five metrics place CoPlayer first on TimingLift and StratEmo.}
Rows: eight frontier models, Base, SFT and CoPlayer.
Darker teal is better within each column; NoiseRate ranks lower as better.
Units: \% except TimingLift (pp) and NoiseRate (per minute).
Local OnTime is n/a because local latency is not comparable to the historical timings.}
\label{fig:result-e}
\end{figure*}

Tables~\ref{tab:reactive} and~\ref{tab:proactive} report the full reactive and proactive results summarized in the main paper; Figure~\ref{fig:result-e} condenses five headline metrics into one rank-colour matrix.

\begin{table*}[t]
\centering
\caption{\textbf{When asked, models read the game but rarely give the right advice in the right tone, and rarely in time.}
Reactive track, 9,103 questions; metrics in Section~\ref{sec:protocol} and Section~\ref{sec:supp-metrics}, in \% except TTFT and TTUA (s).
Frontier models sorted by StratEmo, best last; best in \textbf{bold}, second \underline{underlined} among all eleven models.
CoPlayer is the reactive checkpoint, separate from the proactive checkpoint in Table~\ref{tab:proactive}.
$^\dagger$Served locally; the gray timing entries are HISTORICAL full-benchmark latencies, not the matched serial retest (Appendix~\ref{sec:supp-latency}), and are not ranked.
The reference reply row is a scoring ceiling, not a competing model.}
\label{tab:reactive}
\footnotesize
\setlength{\tabcolsep}{3.345pt}
\begin{tabular}{l ccc ccc ccc cc}
\toprule
& \multicolumn{3}{c}{Game} & \multicolumn{3}{c}{Player} & \multicolumn{3}{c}{Time} & \multicolumn{2}{c}{Rules} \\
\cmidrule(lr){2-4}\cmidrule(lr){5-7}\cmidrule(lr){8-10}\cmidrule(lr){11-12}
Model & GroundF1$\uparrow$ & StratAcc$\uparrow$ & FactAcc$\uparrow$ & EmoAcc$\uparrow$ & SupQual$\uparrow$ & \textbf{StratEmo}$\uparrow$ & TTFT$\downarrow$ & TTUA$\downarrow$ & \textbf{OnTime}$\uparrow$ & Violation$\downarrow$ & LenOK$\uparrow$ \\
\midrule
HY4                  & 34.6 & 24.7 & 70.4 & 47.4 & 36.4 & 5.7 & 5.9 & 7.0 & 10.3 & 25.3 & 84.1 \\
DeepSeek-V4.1-Flash  & 31.4 & 25.5 & 63.7 & 44.6 & 34.5 & 6.4 & \textbf{1.9} & \textbf{2.2} & \textbf{29.4} & 30.6 & 74.0 \\
Doubao-Seed-2.1-Pro  & 34.3 & 21.7 & 70.6 & 57.0 & 37.8 & 6.8 & \underline{3.3} & \underline{4.1} & 15.8 & 25.0 & 96.3 \\
GPT-5.6-Sol          & 42.1 & 29.3 & \underline{78.2} & 51.9 & 40.8 & 7.9 & 5.2 & 5.1 & 17.0 & 21.0 & 90.5 \\
Kimi-K3              & 33.4 & 23.9 & 68.2 & 54.3 & 39.0 & 8.6 & 3.9 & 4.9 & 14.6 & 30.0 & 75.4 \\
Claude-Sonnet-5.5    & 38.6 & 32.8 & 73.5 & 42.3 & 37.8 & 8.8 & 4.8 & 5.5 & 18.1 & 26.9 & 50.9 \\
Qwen3.8-Max          & 34.3 & 23.8 & 69.8 & 67.0 & 44.6 & 12.3 & 7.5 & 8.3 & 10.6 & 27.0 & 82.9 \\
Gemini-3.8-Flash     & 39.7 & 31.1 & 69.0 & 54.8 & 41.1 & 14.5 & 4.7 & 4.8 & \underline{19.6} & 25.9 & 94.0 \\
\midrule
\multicolumn{12}{l}{\emph{Local base and CoPlayer models}} \\
Base                 & 26.7 & 16.0 & 63.4 & 55.8 & 35.1 & 4.7 & \textcolor{gray}{7.4$^\dagger$} & \textcolor{gray}{20.2$^\dagger$} & \textcolor{gray}{0.8$^\dagger$} & 30.9 & 81.9 \\
\rowcolor{coplayerblue}
SFT                  & \underline{48.8} & \underline{38.6} & 67.7 & \underline{83.2} & \underline{57.4} & \underline{29.0} & \textcolor{gray}{5.4$^\dagger$} & \textcolor{gray}{17.8$^\dagger$} & \textcolor{gray}{0.9$^\dagger$} & \underline{20.2} & \underline{98.3} \\
\rowcolor{coplayerblue}
CoPlayer             & \textbf{55.1} & \textbf{46.1} & \textbf{84.5} & \textbf{87.5} & \textbf{61.5} & \textbf{34.3} & \textcolor{gray}{6.4$^\dagger$} & \textcolor{gray}{19.8$^\dagger$} & \textcolor{gray}{0.5$^\dagger$} & \textbf{13.5} & \textbf{99.0} \\
\midrule
Reference reply      & 98.0 & 100.0 & 98.5 & 99.2 & 99.0 & 99.1 & -- & -- & -- & 0.4 & 99.0 \\
\bottomrule
\end{tabular}
\end{table*}

\begin{table*}[t]
\centering
\caption{\textbf{Without being asked, CoPlayer concentrates its lines in required windows beyond a frequency-matched random speaker.}
Proactive track, 818 videos, 4,045 required events; metrics in Section~\ref{sec:protocol} and Section~\ref{sec:supp-metrics}, in \% except TimingLift (percentage points), SpeakRate and NoiseRate (per minute) and TTFT (s); recall intervals are within $\pm$1.6.
Frontier models sorted by TimingLift, best last; best in \textbf{bold}, second \underline{underlined} among all eleven models.
$^\dagger$Latency-dependent and not ranked; gray timing entries are HISTORICAL full-benchmark latencies, not the matched serial retest of Appendix~\ref{sec:supp-latency}.}
\label{tab:proactive}
\footnotesize
\setlength{\tabcolsep}{3.485pt}
\begin{tabular}{l cccc cc ccc ccc c}
\toprule
& \multicolumn{4}{c}{When to speak} & \multicolumn{2}{c}{Observation} & \multicolumn{3}{c}{What to say} & \multicolumn{3}{c}{When to stay silent} & Speed \\
\cmidrule(lr){2-5}\cmidrule(lr){6-7}\cmidrule(lr){8-10}\cmidrule(lr){11-13}\cmidrule(lr){14-14}
Model & Recall$\uparrow$ & Recall-S$\uparrow$ & \makecell{Rand\\Base} & \textbf{TimingLift}$\uparrow$ & \makecell{Seen\\Rate}$\uparrow$ & Recall-O$\uparrow$ & \makecell{Hit\\Strat}$\uparrow$ & \makecell{Hit\\Sup}$\uparrow$ & \makecell{Hit\\StratEmo}$\uparrow$ & \makecell{Speak\\Rate} & \makecell{Noise\\Rate}$\downarrow$ & \makecell{Quota\\Over}$\downarrow$ & TTFT$\downarrow$ \\
\midrule
DeepSeek-V4.1-Flash  & 72.1 & \textbf{68.2} & 60.8 & +11.3 & 74.7 & \textbf{96.6} & 13.2 & 16.9 & 2.4 & 8.8 & 4.98 & 87.6 & \textbf{1.7} \\
Kimi-K3              & 40.6 & 30.1 & 25.9 & +14.7 & 45.0 & 90.3 & 18.3 & 25.6 & 6.6 & 3.4 & 1.76 & 57.8 & 2.8 \\
Qwen3.8-Max          & 57.0 & 47.9 & 40.8 & +16.2 & 59.8 & 95.3 & 13.0 & 20.8 & 3.1 & 5.3 & 3.26 & 82.6 & 2.8 \\
HY4                  & 55.1 & 45.0 & 38.8 & +16.3 & 61.5 & 89.7 & 13.8 & 16.0 & 1.9 & 4.9 & 2.10 & 71.0 & 3.5 \\
Doubao-Seed-2.1-Pro  & \underline{73.9} & \underline{67.6} & 55.4 & +18.6 & 77.3 & \underline{95.7} & 10.4 & 18.1 & 2.3 & 7.7 & 4.72 & 91.9 & \underline{2.0} \\
Claude-Sonnet-5.5    & 66.3 & 56.1 & 45.4 & +20.8 & \underline{79.3} & 83.5 & 16.5 & 21.5 & 2.8 & 5.7 & 2.79 & 75.9 & 3.6 \\
Gemini-3.8-Flash     & 40.5 & 20.5 & 17.8 & +22.7 & 69.0 & 58.6 & 19.8 & 29.8 & 8.0 & 1.5 & \underline{0.64} & \underline{24.1} & 4.1 \\
GPT-5.6-Sol          & \underline{73.9} & 55.6 & 45.9 & +28.0 & \textbf{82.1} & 90.1 & 24.1 & 21.3 & 4.0 & 5.4 & 2.25 & 69.9 & 4.3 \\
\midrule
\multicolumn{14}{l}{\emph{Local base and CoPlayer models}} \\
Base                 & \textbf{84.0} & \textcolor{gray}{80.1$^\dagger$} & 58.4 & +25.6 & \textcolor{gray}{93.3$^\dagger$} & \textcolor{gray}{90.0$^\dagger$} & 5.8 & 18.6 & 1.7 & 8.2 & 5.75 & 84.4 & \textcolor{gray}{0.7$^\dagger$} \\
\rowcolor{coplayerblue}
SFT                  & 55.0 & \textcolor{gray}{45.2$^\dagger$} & 15.2 & \underline{+39.9} & \textcolor{gray}{97.8$^\dagger$} & \textcolor{gray}{56.3$^\dagger$} & \underline{36.5} & \underline{42.7} & \underline{24.0} & 1.1 & \textbf{0.45} & \textbf{16.6} & \textcolor{gray}{2.3$^\dagger$} \\
\rowcolor{coplayerblue}
CoPlayer             & 67.4 & \textcolor{gray}{57.3$^\dagger$} & 23.7 & \textbf{+43.7} & \textcolor{gray}{97.8$^\dagger$} & \textcolor{gray}{68.9$^\dagger$} & \textbf{48.2} & \textbf{45.2} & \textbf{31.2} & 1.9 & 0.97 & 33.1 & \textcolor{gray}{2.1$^\dagger$} \\
\bottomrule
\end{tabular}
\end{table*}

Table~\ref{tab:bias} removes the events of each of the two annotators with enough removed events.
The reactive lead survives: on the 2,310 advice questions not annotated by Claude-Opus-5.5, CoPlayer reaches a StratEmo of 28.4 against 10.5 for the best frontier model on those same questions.
Table~\ref{tab:split} separates in-play from aftermath questions and card and board games from visual games; every model is less timely on the in-play questions, \mbox{whose deadline is at most 5 seconds.}

\begin{table}[ht]
\centering
\caption{\textbf{Rankings hold when the annotators' own events are removed.}
Reactive StratAcc and StratEmo (\%) on all advice questions, on those not annotated by Claude-Opus-5.5 (w/o Opus, 2,310 questions) and on those not annotated \mbox{by GPT-6.1-Sol (w/o GPT, 5,379 questions).}}
\label{tab:bias}
\footnotesize
\setlength{\tabcolsep}{2.73pt}
\begin{tabular}{l ccc ccc}
\toprule
& \multicolumn{3}{c}{StratAcc$\uparrow$} & \multicolumn{3}{c}{StratEmo$\uparrow$} \\
\cmidrule(lr){2-4}\cmidrule(lr){5-7}
Model & All & w/o Opus & w/o GPT & All & w/o Opus & w/o GPT \\
\midrule
HY4                  & 24.7 & 23.9 & 24.8 & 5.7 & 6.1 & 5.2 \\
\makecell[l]{DeepSeek-\\V4.1-Flash} & 25.5 & 24.2 & 26.0 & 6.4 & 5.8 & 6.3 \\
\makecell[l]{Doubao-\\Seed-2.1-Pro} & 21.7 & 19.0 & 22.8 & 6.8 & 5.9 & 7.2 \\
GPT-5.6-Sol          & 29.3 & 28.7 & 29.2 & 7.9 & 7.8 & 7.5 \\
Kimi-K3              & 23.9 & 20.8 & 25.2 & 8.6 & 6.7 & 9.3 \\
\makecell[l]{Claude-\\Sonnet-5.5} & 32.8 & 28.2 & 33.9 & 8.8 & 7.5 & 8.8 \\
Qwen3.8-Max          & 23.8 & 19.7 & 25.5 & 12.3 & 8.9 & 13.6 \\
\makecell[l]{Gemini-\\3.8-Flash} & 31.1 & 26.9 & 32.5 & 14.5 & 10.5 & 15.7 \\
\midrule
\rowcolor{coplayerblue}
SFT                  & 38.6 & 31.9 & 39.9 & 29.0 & 21.4 & 30.7 \\
\rowcolor{coplayerblue}
CoPlayer             & 46.1 & 42.6 & 45.9 & 34.3 & 28.4 & 34.8 \\
\bottomrule
\end{tabular}
\end{table}

\begin{table}[ht]
\centering
\caption{\textbf{Fast decisions are where every model fails to be on time.}
StratEmo and OnTime (\%) on in-play versus aftermath questions, and on card and board games versus visual games.}
\label{tab:split}
\footnotesize
\setlength{\tabcolsep}{2.611pt}
\begin{tabular}{l cc cc cc cc}
\toprule
& \multicolumn{2}{c}{\makecell{StratEmo\\by question}} & \multicolumn{2}{c}{\makecell{OnTime\\by question}} & \multicolumn{2}{c}{\makecell{StratEmo\\by game}} & \multicolumn{2}{c}{\makecell{OnTime\\by game}} \\
\cmidrule(lr){2-3}\cmidrule(lr){4-5}\cmidrule(lr){6-7}\cmidrule(lr){8-9}
Model & \makecell{In-\\play} & After & \makecell{In-\\play} & After & Board & Visual & Board & Visual \\
\midrule
HY4                  & 5.9 & 5.3 & 0.1 & 23.6 & 6.4 & 5.1 & 0.3 & 16.3 \\
\makecell[l]{DeepSeek-\\V4.1-Flash} & 6.5 & 6.1 & 9.6 & 55.2 & 6.9 & 6.0 & 15.6 & 37.7 \\
\makecell[l]{Doubao-\\Seed-2.1-Pro} & 8.0 & 4.7 & 2.4 & 33.3 & 8.1 & 5.9 & 4.3 & 22.8 \\
GPT-5.6-Sol          & 8.7 & 6.4 & 1.3 & 37.5 & 9.9 & 6.4 & 3.3 & 25.3 \\
Kimi-K3              & 9.0 & 7.9 & 1.2 & 31.9 & 9.5 & 7.9 & 3.8 & 21.1 \\
\makecell[l]{Claude-\\Sonnet-5.5} & 8.5 & 9.3 & 1.4 & 39.9 & 10.1 & 7.8 & 3.3 & 27.1 \\
Qwen3.8-Max          & 12.6 & 11.8 & 0.8 & 23.3 & 15.9 & 9.6 & 1.9 & 15.8 \\
\makecell[l]{Gemini-\\3.8-Flash} & 16.7 & 10.3 & 1.9 & 42.8 & 20.2 & 10.1 & 4.9 & 28.5 \\
\bottomrule
\end{tabular}
\end{table}

\begin{figure*}[ht]
\centering
\begin{minipage}[b]{0.465\textwidth}\centering
\includegraphics[width=\linewidth]{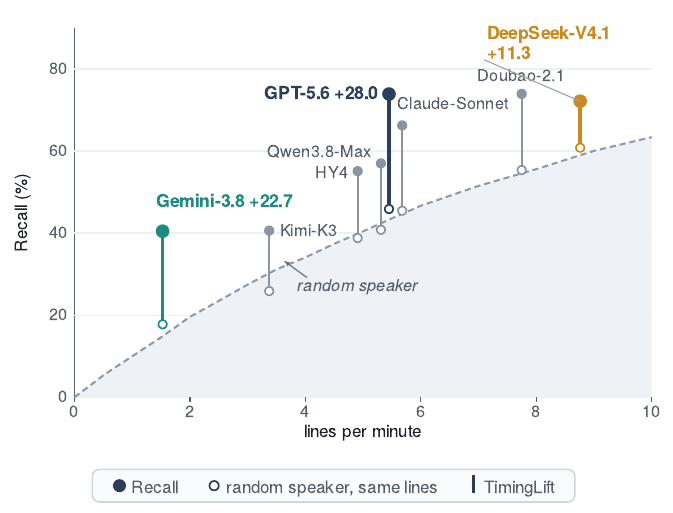}\\[-1pt]
{\small (a) TimingLift}
\end{minipage}\hfill
\begin{minipage}[b]{0.505\textwidth}\centering
\includegraphics[width=\linewidth]{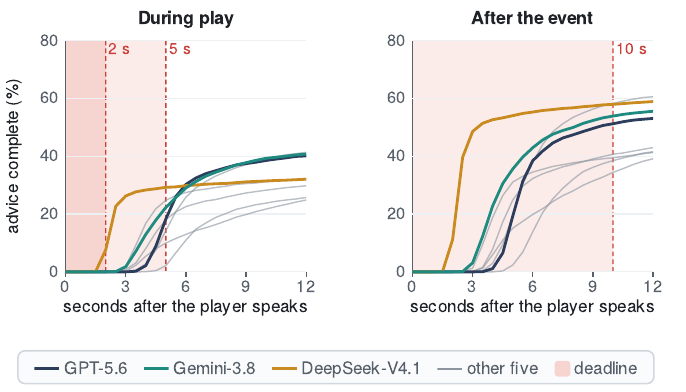}\\[-1pt]
{\small (b) Time until the advice is complete}
\end{minipage}
\caption{\textbf{Raw recall rewards talking, and advice asked during play arrives too late.}
(a) Each model's Recall (filled) against RandBase, a random speaker with as many lines (open); the bar between them is TimingLift.
GPT-5.6-Sol and Doubao-Seed-2.1-Pro reach the same recall, but the latter talks over 40\% more.
(b) Share of reactive questions whose advice is complete by each second; shading ends at the OnTime deadlines (2\,s for 85\% of in-play questions), and many replies never contain complete advice.}
\label{fig:results}
\end{figure*}

Raw recall rewards talking, and advice asked during play arrives too late; Figure~\ref{fig:results} shows both effects.
GPT-5.6-Sol and Doubao-Seed-2.1-Pro reach the same recall, but Doubao speaks more than 40\% more often, and many replies never contain complete advice before the deadline.
DeepSeek-V4.1-Flash is the most timely frontier model (OnTime 29.4\%), and Gemini-3.8-Flash is the quietest (NoiseRate 0.64, QuotaOver 24.1\%) but misses most required windows (Recall 40.5\%), so quietness alone does not make a companion well timed.
A self-consistency check of the benchmark judge reproduces 94.3\% of its StratAcc verdicts and 97.7\% of its support scores within one point on the five-point support scale.

\begin{figure*}[t]
\centering
\includegraphics[width=\textwidth]{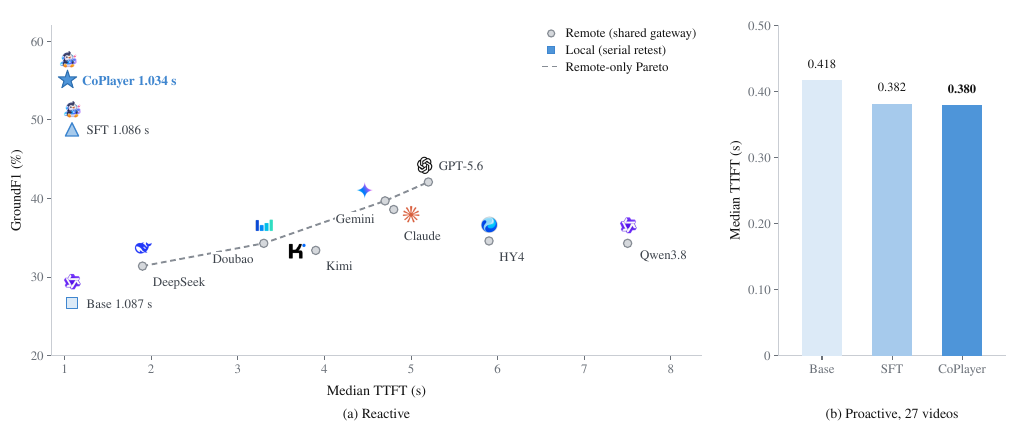}
\caption{\textbf{Training improves grounding with similar single-request latency.}
(a) Full-benchmark GroundF1 versus median TTFT for eight remote models and three local checkpoints; the dashed line marks the remote-only Pareto frontier.
Remote timings are historical; local timings are the 180-question serial retest with one request in flight on two H20 GPUs.
(b) Per-call TTFT over 27 proactive videos.
Quality and local latency summarize different samples; remote and local timings reflect different deployments.}
\label{fig:pareto}
\end{figure*}

Figure~\ref{fig:pareto} compares full-benchmark grounding quality with observed deployment latency and reports local proactive latency alongside the reactive comparison.
The serial retest served Base, SFT and CoPlayer one at a time on one TP2 replica using two H20 GPUs, with one request in flight globally.
It contains 180 reactive questions (20 per game; 60 urgent, 30 strategic and 90 emotional; 139 advice questions) and 27 complete proactive videos (239.4\,min) after 20 serial warmup requests per stage.
Reactively, CoPlayer's median TTFT is 1.034\,s versus 1.086\,s for SFT and 1.087\,s for Base; it returns locatable advice on 69 rather than 59 of 139 questions and lowers the no-valid-advice rate from 57.6\% to 50.4\%.
Proactively, Base, SFT and CoPlayer have per-call medians of 0.418, 0.382 and 0.380\,s over 3,618, 3,634 and 3,634 calls, respectively.
TTFT includes queueing, prefill and self-generated context, so these are end-to-end deployment differences; each checkpoint is one uninterrupted run, and we draw no error bars or significance claim.

\subsection{Serial latency retest}
\label{sec:supp-latency}

\paragraph{Inputs and concurrency.}
On the reactive track each of the 180 questions is one turn with four 2$\times$2 grids tiling the last 60\,s (16 source frames at the recorded frame times), followed by the player's utterance; the 720 grid images and 2,880 frame slots are the same across checkpoints.
The proactive track uses 27 complete videos, three per game, totaling 239.4\,min, with a 1\,fps cache of about 14.4k frames.
At every call the model receives one 2$\times$2 grid over the last step (four cached frames; steps are 2, 4 or 6\,s by game), its own memo, its last eight spoken lines and the fixed opening instruction, so active context is self-generated and differs across checkpoints.
Base, SFT and CoPlayer were served one at a time as a single TP2 replica on two H20 GPUs (GPU0 and GPU1); the other six GPUs were intentionally unused.
The server ran with max-num-seqs 1, 8,192 context, at most 32,768 batched tokens, thinking disabled, temperature 0.3 and a 1,024-token output cap, and the client kept one request in flight globally.
Each stage used the same 20 serial warmup requests, excluded from the 540 reactive responses and 81 proactive traces.

\paragraph{Measurement boundary.}
TTFT runs from just before the HTTP request to the first nonempty visible content; it includes request handling, server queueing, prompt computation and the first output step, but not image preparation, model loading, warmup or client-side limiter waits.
The client reads 512-byte SSE chunks, so content may arrive buffered; TTFT is therefore a first-visible-client-content time, not pure server compute.
Proactive TTFT is the first visible character of the JSON reply, not necessarily the first spoken word.
Reactive medians use the lower order statistic $x_{\lfloor q(n-1)\rfloor}$ at $q=0.5$; proactive per-call medians are ordinary medians of the per-call values.
The proactive medians pool 3,618/3,634/3,634 calls from the 27 videos and are per call, not video-level medians; the counts differ because each reply advances stream time by its generation latency.
TTUA locates the judge-extracted end of the advice substring in the reply's cumulative character timestamps and is computed only where that substring is locatable; the counts are 31, 59 and 69 of 139, so the three medians condition on different subsets and are not full-set success times.
Missing advice counts as an OnTime failure and is reported separately.
Strict recall counts a required event only when an utterance time plus that call's TTFT falls inside the window, whereas ideal recall ignores the offset.
Each checkpoint is one uninterrupted run without repeat trials, confidence intervals or significance tests, and the sample is 180 questions and 27 videos rather than the full 9,103-question reactive benchmark and its 818 proactive videos.

\paragraph{Historical full-benchmark timings.}
The latencies in Tables~\ref{tab:reactive} and~\ref{tab:proactive} predate the serial retest and remain historical.
Base and SFT were served together as eight TP=2 replicas each on four 8-GPU hosts (16 GPUs per checkpoint), while the reactive CoPlayer policy ran separately as 12 TP=2 replicas on three 8-GPU hosts (24 GPUs).
The pooled base median combines 4,191 records from an earlier three-replica single-host run (median 14.893\,s) with 4,912 records from the final run (median 5.221\,s); SFT's 5.414\,s comes entirely from the final run.
On the 4,912 questions served in the same final run, the matched SFT$-$base median difference is $-0.066$\,s (95\% bootstrap CI $-0.217$ to $+0.012$; SFT faster on 51.3\%), so this later subset shows no clear TTFT advantage for either checkpoint, and a confidence interval containing zero is not proof of equivalence.
The run manifest marks its baselines as not latency-matched, so its 6.410\,s cannot be read as a model-speed difference, and these timings are not pooled with the serial retest.
Each historical reactive prompt carries four 2$\times$2 grids from the preceding 60\,s (median 4,116 prompt tokens locally), and each replica received up to 24 concurrent requests.
These historical runs used vLLM 0.28.0 with BF16 weights and no quantization, tensor parallelism 2, 8,192 context, 16,384 batched tokens per replica, temperature 0.3 and a 1,024-token output cap.
Run logs report 95.0\,GiB of device memory per GPU but no model string; the H20 identification comes from the project's deployment notes, not from hardware telemetry.

\paragraph{Earlier checkpoints.}
An earlier single-policy RL run trained both tracks together and reached a reactive StratEmo of 36.5, but with a lower FactAcc (81.8) and a higher Violation (14.1) than the reactive checkpoint reported here; its proactive recall was 47.6.
A later proactive-only run reached a Recall of 73.0 and a TimingLift of +43.3, but HitStratEmo fell to 11.4 and NoiseRate rose to 1.77.
Neither is the track-specific model reported in the main tables.

\subsection{Resource-interference control}
\label{sec:supp-cohost}

A separate cohost trial checks whether sharing the host changes the latency picture.
All four checkpoints stayed resident, each on its own H20 pair (8 GPUs in total), with at most one request per model and up to four concurrent requests across the host.
The phases differ only in activity: the resident-serial phase issues one request at a time, while the cohost phase issues one request to each model; all four weights are resident in both phases.
The trial uses 36 reactive questions and 27 proactive snapshots frozen from a prior SFT trajectory, not complete videos, and reports the reactive checkpoint on the reactive questions and the proactive checkpoint on the snapshots.
Table~\ref{tab:cohost} gives the six track-matched rows; the other RL checkpoint was resident as a cross-track control but is not a core reported model.
Reactive cohost medians are 0.197, 0.249 and 0.217\,s for Base, SFT and CoPlayer, within 0.021\,s of their resident-serial medians in this trial; the frozen snapshots are faster in cohost (0.180--0.183 versus 0.356--0.364\,s), which is a resource-interference observation, not evidence of intrinsic acceleration or significance.
Phase order was alternated by case, but prefix caching was not fully controlled, and both phases use the same 512-byte SSE reader.
These short-trial values are not pooled with the 180-question serial Pareto; they serve only as a resource-interference control for the serial comparison.

\begin{table}[ht]
\centering
\caption{\textbf{Co-hosting shifts reactive median TTFT by at most 0.021\,s.}
Four checkpoints stay resident on separate H20 pairs; each phase has 36 reactive questions and 27 frozen proactive snapshots.
TTFT is in seconds; paired $\Delta$ is the per-case median difference, cohost minus resident serial, on matched inputs.}
\label{tab:cohost}
\footnotesize
\setlength{\tabcolsep}{10.8pt}
\begin{tabular}{l c c c c}
\toprule
Model & \makecell{Resident\\serial} & Cohost & \makecell{Cohost\\P90} & \makecell{Paired\\$\Delta$} \\
\midrule
\multicolumn{5}{l}{\emph{Reactive, 36 questions}} \\
Base     & 0.194 & 0.197 & 1.100 & $+0.009$ \\
\rowcolor{coplayerblue}
SFT      & 0.255 & 0.249 & 1.126 & $-0.001$ \\
\rowcolor{coplayerblue}
CoPlayer & 0.196 & 0.217 & 1.119 & $+0.021$ \\
\midrule
\multicolumn{5}{l}{\emph{Proactive snapshots, 27 frozen}} \\
Base     & 0.361 & 0.180 & 0.394 & $-0.211$ \\
\rowcolor{coplayerblue}
SFT      & 0.356 & 0.182 & 0.394 & $-0.205$ \\
\rowcolor{coplayerblue}
CoPlayer & 0.364 & 0.183 & 0.400 & $-0.212$ \\
\bottomrule
\end{tabular}
\end{table}

\section{Metric details}
\label{sec:supp-metrics}

\subsection{Protocol details}
\label{sec:supp-protocol}

\noindent\textbf{Reactive counts.}
The 5,153 Bench events yield 5,153 in-play questions and 3,950 aftermath questions, 9,103 in total; 7,037 are advice questions.
Of the in-play questions, 4,403 (85.4\%) carry the 2\,s urgent deadline, while other in-play questions use 5\,s and aftermath questions 10\,s.
Each question shows 16 frames uniformly sampled from the preceding 60\,s in four 2$\times$2 grids.

\noindent\textbf{Proactive counts.}
The timing audit leaves 5,133 of 5,153 events for evaluation: 4,045 required and 1,088 optional.
The step $\Delta t$ is 2\,s for the four card and board games, 4\,s for Honor of Kings, Plants vs.\ Zombies 2, Carrot Fantasy and Hearthstone, and 6\,s for Stardew Valley.
Each step observes the previous $\Delta t$ seconds with the memo (at most 25 words) and the last eight lines.
The quiet-time rule allows one chat line per 2-minute period, or 1 minute for Stardew Valley, and RandBase uses 20 uniform draws per video.

\begin{table*}[ht]
\centering
\caption{\textbf{Metric abbreviations used in Tables~\ref{tab:reactive} and~\ref{tab:proactive}, with definitions for every reported metric.}}
\label{tab:acronyms}
\footnotesize
\setlength{\tabcolsep}{18.16pt}
\begin{tabular}{l l l l}
\toprule
Acronym & Meaning & Acronym & Meaning \\
\midrule
GroundF1 & must-mention F1, wrong claims as false positives & StratAcc & advice questions with sound strategy \\
FactAcc & verifiable facts correct & EmoAcc & player emotion recognized \\
SupQual & support score 0--4, rescaled & StratEmo & sound strategy with support $\geq 3$ \\
TTFT & time to first visible token & TTUA & time until useful advice is complete \\
OnTime & advice complete by its deadline (TTUA) & OnTime-F & OnTime under a fixed deadline \\
Violation & reply violates a must-not item & LenOK & reply within the word limit \\
Recall & required windows hit at step time & Recall-S & hit by first-token arrival \\
RandBase & random speaker, same utterance count & TimingLift & Recall minus RandBase \\
SeenRate & required windows with an observed step & Recall-O & recall on observed windows \\
HitStrat & sound strategy in first in-window line & HitSup & support score of that line \\
HitStratEmo & sound strategy and support $\geq 3$ & SpeakRate & lines per minute \\
NoiseRate & outside-window chat, wrong or noise per minute & QuotaOver & quiet-time periods over quota \\
\bottomrule
\end{tabular}
\end{table*}

\paragraph{Reactive track.}
For every reply the judge marks which must-mention points it covers, extracts its claims about the game and labels each as supporting the ground truth, irrelevant or wrong, extracts its verifiable facts and labels each as correct, wrong or unverifiable, decides whether its main advice is a sound strategy and whether it perceives the player's emotion, and rates its emotional support from 0 to 4.
With $\mathrm{TP}$ the covered points, $\mathrm{FN}$ the missed points and $\mathrm{FP}$ the wrong claims, summed over all questions, GroundF1 is the F1 of $P = \mathrm{TP}/(\mathrm{TP}+\mathrm{FP})$ and $R = \mathrm{TP}/(\mathrm{TP}+\mathrm{FN})$.
A question is an advice question when the judge rates the reference reply itself as containing sound advice; a reply without advice fails StratAcc.
FactAcc ignores unverifiable facts.
SupQual is the mean support score divided by 4: 0 for a wrong or mocking response, 1 for template comfort, 2 for support not tied to the game, 3 for support tied to the specific situation, and 4 for support that also matches the expected tone.
A companion must get the strategy and the emotion right in the same reply, so
\begin{equation}
\mathrm{StratEmo} = \frac{1}{|\mathcal{Q}_{\mathrm{adv}}|}\sum_{q \in \mathcal{Q}_{\mathrm{adv}}} \mathbb{1}\big[\mathrm{strategy}_q \wedge s_q \geq 3\big]
\label{eq:supp-stratemo}
\end{equation}
counts the advice questions $\mathcal{Q}_{\mathrm{adv}}$ whose reply gives sound advice with fitting support; $\mathrm{strategy}_q$ is its binary verdict for question $q$.
For TTUA, the judge quotes the last words of the first complete piece of advice, and TTUA is the time at which the streamed reply first contains them.
Speed matters only relative to the moment, so OnTime counts a reply as timely when its advice is complete before
\begin{equation}
D = \min\big(d_s, \max(0.5\,\mathrm{s}, w)\big),
\label{eq:supp-deadline}
\end{equation}
where $w$ is the time left in the event's window and $d_s$ is 2\,s for urgent in-play alerts, 5\,s for other in-play questions and 10\,s after the event.
Of the 5,153 in-play questions, 4,403 (85.4\%) carry the 2\,s deadline.
OnTime-F uses the fixed deadline $D = d_s$ without the window bound; Violation is the share of replies that violate a must-not item, and LenOK the share within the word limit.
All reactive metrics other than TTFT and TTUA are reported in percent.

\paragraph{Proactive track.}
At every step the model receives its memo from the previous step and its last eight lines, and returns a JSON object with a speak flag, the line and an updated memo of at most 25 words.
Recall uses the step time (ideal clock) and Recall-S the time the first token reaches the player (strict clock).
HitStrat and HitStratEmo exclude events whose type calls only for emotion, such as the end of a game or a big harvest.
A period counts toward QuotaOver when it contains more than one chat, wrong or noise line outside the windows; periods last two minutes for most games, and one minute for Stardew Valley.
Equation~\ref{eq:tg} defines TimingLift; the random speaker places $|U_v|$ utterances uniformly over the covered stream, and RandBase is estimated with 20 draws per video.

\paragraph{Hidden reasoning.}
With reasoning switches off, hidden reasoning still appears reactively on 0.2\%, 4.2\% and 52.5\% of calls for Kimi-K3, Claude-Sonnet-5.5 and Gemini-3.8-Flash (101.9 tokens on average), and proactively on 20.5\%, 0.1\% and 72.9\%; the other five models emit none.

\section{Rewards of the RL policies}
\label{sec:supp-reward}

Shared supervision used 1,255 Train videos, giving 12,914 reactive question--answer examples and 27,843 step-level proactive examples.
Both policies start from that supervised checkpoint and are trained with GRPO on the 302 Train videos that supervised fine-tuning did not use, with 26 further videos held out for validation.
Each step samples eight responses per prompt at temperature 1.0, and a locally served Qwen3.8-Flash-Next applies the judge prompts of the corresponding track to every sampled response.
This reward judge uses the same prompts and criteria as the benchmark judge Qwen3.8-Flash, but it is a different checkpoint: on 2,000 paired samples the two agree on 93.3\% of strategy verdicts, 84.6\% of emotion verdicts and 98.7\% of support scores within one point of each other on the five-point support scale.
We train for 200 steps with the vision encoder frozen and learning rate $10^{-6}$, and we zero the advantage of a group whose rewards have a standard deviation below 0.01.
The reactive policy uses batches of 24 questions and a KL coefficient of 0.005; the proactive policy uses batches of 40 steps and 0.01.
The two policies are separate checkpoints trained from the shared initialization; they are not one model trained jointly on both tracks.

\paragraph{Reactive reward: what a good reply is worth.}
Every sampled reply is scored by the judge, and the reward is
\begin{equation}
r = \mathrm{clip}\big(0.15\,r_{\mathrm{fmt}} + 0.55\,r_{\mathrm{con}} + 0.35\,r_{\mathrm{emo}} - p,\; -0.5,\; 1\big).
\label{eq:supp-reward}
\end{equation}
Content carries the largest weight because a companion that reads the game wrongly is worse than one that says nothing, and the emotion term is close behind it because a player who asks for help while losing is also asking to be taken seriously.
Format carries the least, since it governs how a reply is packaged rather than whether it helps.

The content term $r_{\mathrm{con}} = 0.45\,c + 0.35\,a + 0.20\,f$ combines the coverage $c$ of the must-mention points, sound advice $a \in \{0,1\}$ and the share $f$ of correct facts among the verifiable ones; $c$ and $f$ default to 0.6 when undefined, and on a question without an advice verdict the weights of $c$ and $f$ are rescaled to sum to one.
Facts weigh least of the three because stating what is on screen is the part the models already do well, so paying more for it would buy description rather than the judgment a companion needs.

The emotion term $r_{\mathrm{emo}} = 0.35\,e + 0.65\,s/4$ combines emotion perception $e \in \{0,1\}$ and the support score $s$.
Support is worth almost twice perception because naming how the player feels costs a clause, while comfort that refers to what just happened on screen is the part players cannot get from a template, and it is the half of StratEmo the models miss most often.
The format term $r_{\mathrm{fmt}} = 0.4\,b + 0.35\,\ell + 0.25\,o$ rewards a well-formed reply $b$, the word limit $\ell$ (1 within the limit, 0.5 up to 1.5 times it, 0 beyond) and advice placed early $o$ (1 when the advice ends in the first half of the reply, falling linearly to 0.5 at its end, and 0.5 when no advice is found).
The penalty $p = 0.5\,v + 0.3\min(1, m/2) + 0.2\,n$ counts a must-not violation $v$, the number $m$ of wrong claims and a quoted engine value $n$ such as a win rate; any reply with a violation, a wrong claim or fact, or an engine value is capped at $-0.4$, and an empty reply receives $-0.5$.

\paragraph{Proactive reward: what a good moment is worth.}
Here the model returns a JSON object with a speak flag, the line and a memo, so the reward scores the decision first and the line second (Table~\ref{tab:reward-active}).
A step whose JSON is malformed, or whose flag disagrees with the line, receives $-1$.
Inside a required window, silence receives $-0.80$, which is below chat and noise but not below malformed JSON, and a wrong or missing required advice receives $-0.55$.
The ordering is deliberate: a companion that misses the teamfight it was watching for has failed at the thing it is there to do, while a clumsy remark during a quiet stretch is merely noise the player can ignore.
Any reward that pays for silence invites a model to buy safety by saying nothing, and a silent companion keeps none of the value a companion is for.

A line the judge rates as sound advice earns $0.35 + 0.05\,c$, and one that also carries situated support ($s \geq 3$) earns $0.85 + 0.05\,c + 0.05\,[s{=}4]$.
The jump between the two is larger than either step alone, because advice and support arriving together are what a player experiences as being accompanied, whereas advice alone reads as a tooltip.
A required event that needs no advice earns $-0.10 + 0.25\,c + 0.65\,s/4$.
Content $c$ enters only as a small term, so a fluent line cannot stand in for a correct call.
Outside every window a useful line is still worth $0.35$, since a companion may volunteer something helpful between key moments, and chat, noise and wrong lines fall to $-0.35$, $-0.5$ and $-0.8$. Outside required windows, a line repeating an earlier one verbatim is paid $-0.6$, because repetition is how a talkative model fills silence without adding anything.
As in the reactive reward, a must-not violation, a wrong claim or fact, or a quoted engine value caps the step at $-0.8$, and the reward is clipped to $[-1, 1]$ before the advantage is computed.

We choose this decision-first profile from four predefined reward profiles on a fixed 720-sample RL-dev proxy; only the \emph{balanced} and \emph{recall} profiles completed 20-step trials, and the \emph{recall} profile was promoted to the 200-step run reported here.
A separate quality-first profile failed its validation gate at step 20 and is not used.

One detail of GRPO needs care here.
A group in which all eight samples stay silent inside a required window carries no gradient, because advantages are normalized within the group, so the miss penalty vanishes exactly where it is needed.
We therefore give the tokens of the \emph{false} decision, and no other token, an advantage of $-0.3$ in such a group.

\begin{table}[ht]
\centering
\caption{\textbf{The proactive reward scores the decision first, then the line.}
Reward of one step; $c$ is the must-mention coverage of Equation~\ref{eq:supp-reward} and $s$ the support score from 0 to 4.}
\label{tab:reward-active}
\footnotesize
\setlength{\tabcolsep}{5.14pt}
\begin{tabular}{l r}
\toprule
Outcome of the step & Reward \\
\midrule
Malformed JSON or flag disagreement & $-1$ \\
\midrule
\emph{Inside a required window} \\
\quad silent & $-0.8$ \\
\quad advice wrong or missing & $-0.55$ \\
\quad sound advice, $s < 3$ & $0.35 + 0.05\,c$ \\
\quad sound advice, $s \geq 3$ & $0.85 + 0.05\,c + 0.05\,[s{=}4]$ \\
\quad no advice needed & $-0.10 + 0.25\,c + 0.65\,s/4$ \\
\midrule
\emph{Outside every window} \\
\quad silent & $0.10$ \\
\quad useful line & $0.35$ \\
\quad chat & $-0.35$ \\
\quad noise / wrong & $-0.5$ / $-0.8$ \\
\quad verbatim repeat & $-0.6$ \\
\bottomrule
\end{tabular}
\end{table}

\section{Data construction details}
\label{sec:supp-data}

We align the events of the three annotators in time, treating two events as the same when their spans overlap with an IoU of at least 0.3, when the shorter span is at least half covered, or when both refer to the same move of a card or board game.
Of the 21,494 model-annotated events, 15,689 (73\%) meet the two-model agreement criterion before video exclusions and human review. Three master's-level human annotators then review the candidates, retaining only those unanimously deemed appropriate.
The retained Train and Bench splits contain 15,201 of them, and in the Bench split all three agree on 74\% of the events.
For each retained event we keep the full annotation of one representative annotator, the agreeing model whose span lies closest to the median span; Claude-Opus-5.5 is the representative for 68\% of Bench events.
All ground truth is in English, translated from Chinese by two models with a third choosing the better version and a 162-entry glossary fixing game terms; we hand-checked a random sample of 30 translated nodes for translation accuracy, not event timing or content.
The timing audit shifted 430 windows (285 of them required), dropped 20 events (13 required) and removed 13 must-mention points; 198 in-play and aftermath questions (2.2\%) still contain a reference reply that refers to the future, and 168 more are uncertain.
Although event annotations undergo human verification, subjective judgments and residual errors may remain. Response-level judge consistency is separately audited in Appendix~\ref{sec:supp-human}.

\section{Case studies}
\label{sec:supp-cases}

\begin{figure*}[p]
\centering
\includegraphics[width=\textwidth]{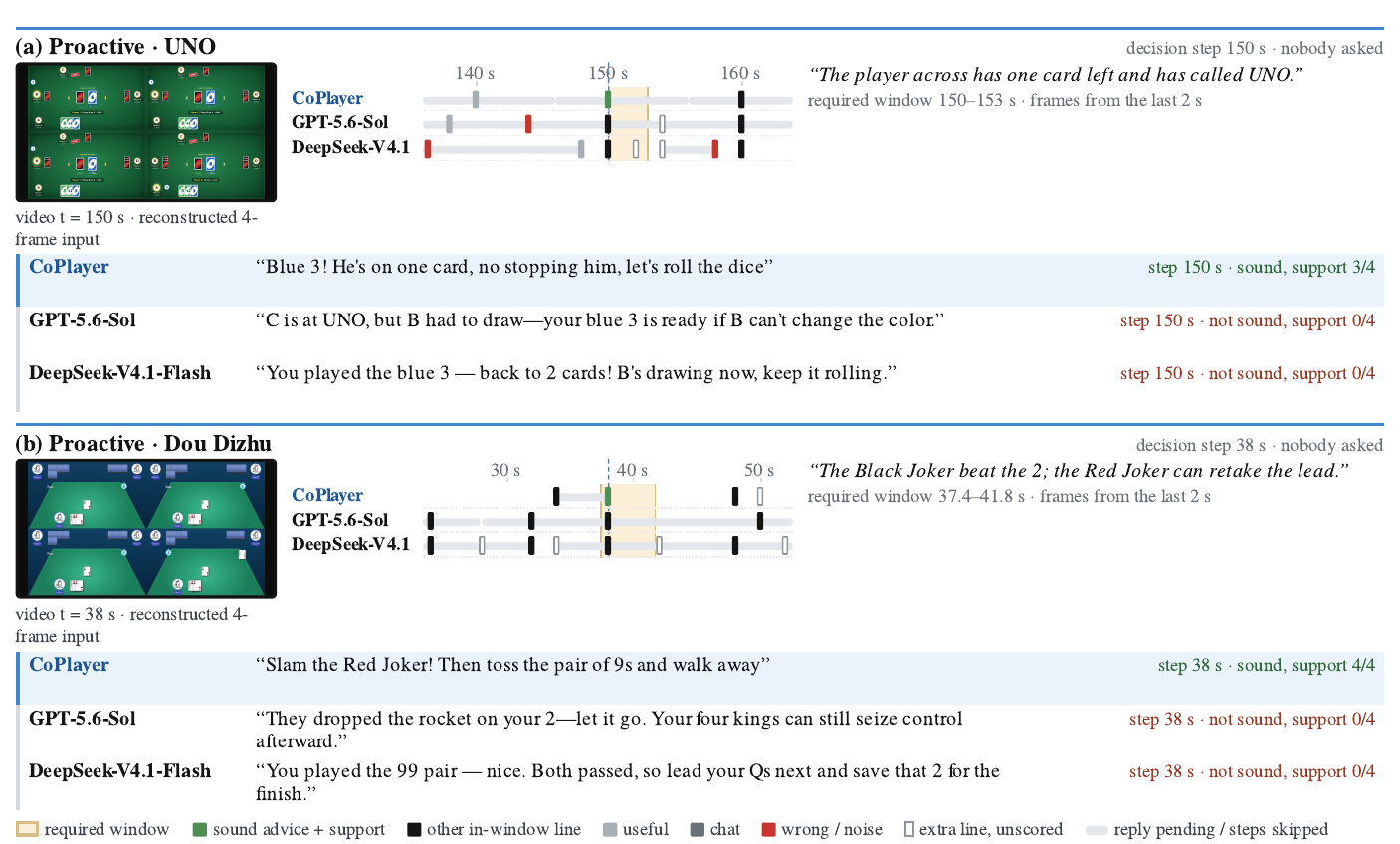}
\caption{\textbf{More proactive interventions on matched inputs.}
(a) UNO and (b) Dou Dizhu events where nobody asked; frames are reconstructed four-frame inputs at the recorded decision step.
Each timeline shows every model's spoken steps and the required window.
CoPlayer gives sound, supported advice in both; GPT-5.6-Sol mentions the playable Blue 3 only conditionally and invents a rocket in Dou Dizhu, while DeepSeek-V4.1-Flash says the Blue 3 was already played and invents a prior play.}
\label{fig:case-proactive}
\end{figure*}

The reactive comparisons below are selected qualitative examples from the final checkpoints, not population estimates.
Each uses the same question and the same sampled model input for every model, and all quotes are verbatim except where an emoji or a trailing excerpt is omitted.
Reactive frames are exact cached model inputs.
Proactive frames in Figure~\ref{fig:case-proactive} are reconstructed four-frame inputs: the original per-step cache was unavailable, so they were rebuilt from SHA256-verified source videos with the runner's recorded sampling rule.

\begin{figure*}[p]
\centering
\includegraphics[width=\textwidth]{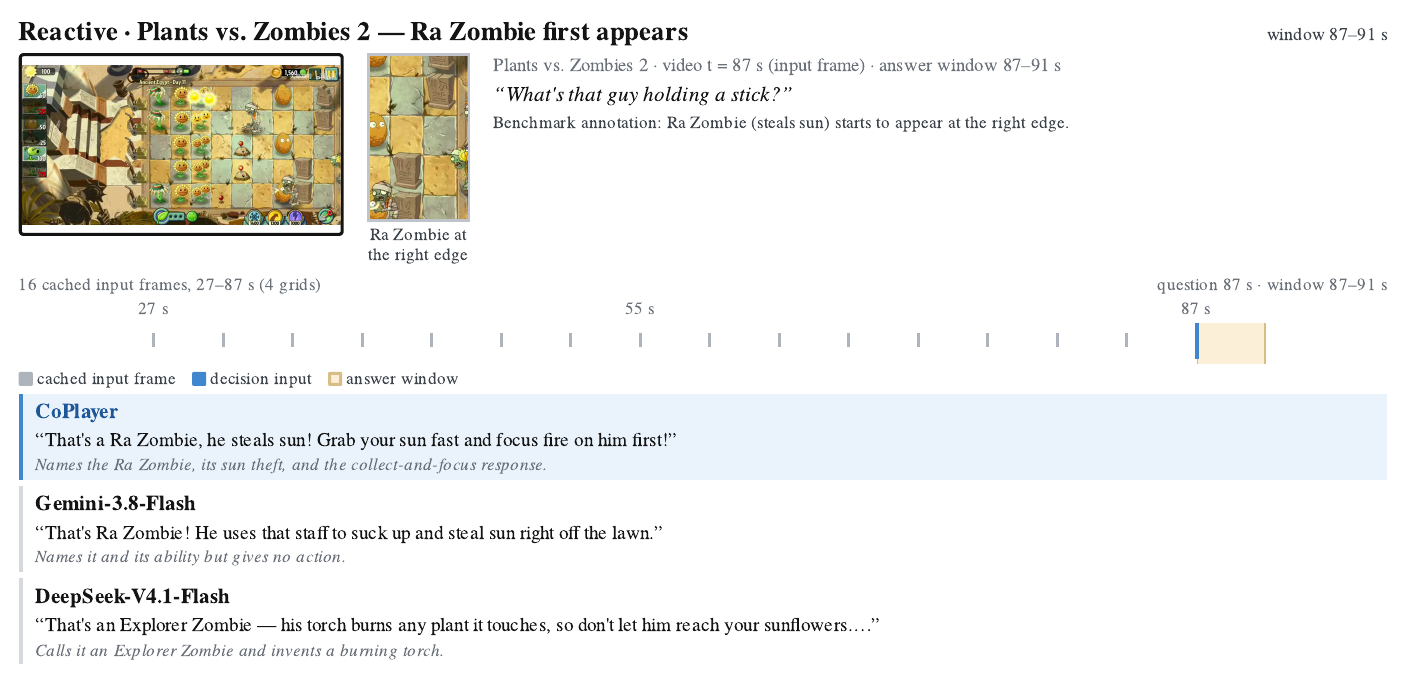}
\caption{\textbf{Plants vs.\ Zombies 2: a Ra Zombie enters.}
The frame is the exact model input at the decision step; the Ra Zombie is entering at the right edge of that input.
CoPlayer names it, explains that it steals sun, and calls for fast sun collection and focus fire; Gemini-3.8-Flash names it but gives no action, and DeepSeek-V4.1-Flash calls it an Explorer Zombie and invents a burning torch.
The timeline shows the 16 cached model-input frames in four grids and the recorded answer window.}
\label{fig:case}
\end{figure*}

\begin{figure*}[p]
\centering
\includegraphics[width=\textwidth]{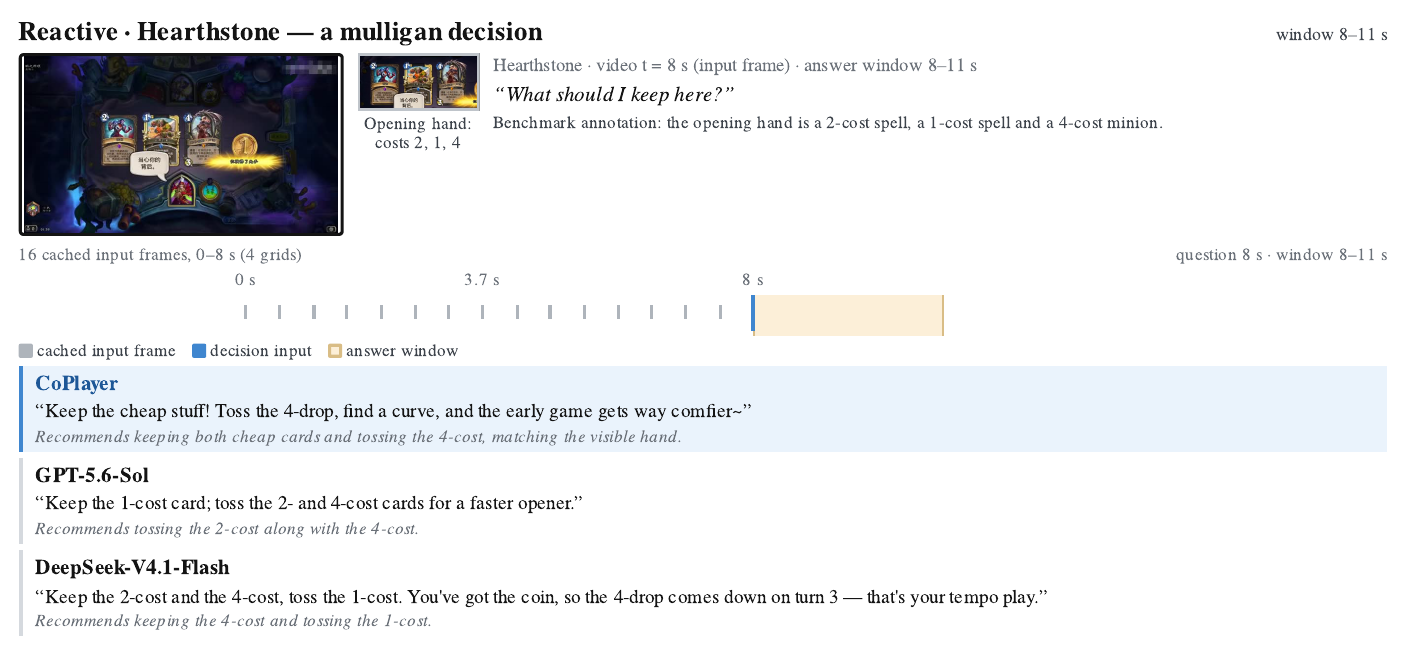}
\caption{\textbf{Hearthstone: a mulligan decision.}
The frame and inset show the opening hand, with a 2-cost, 1-cost and 4-cost card.
CoPlayer keeps both cheap cards and tosses the 4-drop; GPT-5.6-Sol also tosses the 2-cost, while DeepSeek-V4.1-Flash keeps the 4-drop and tosses the 1-cost.
The timeline shows the 16 cached frames and answer window.}
\label{fig:case-hs}
\end{figure*}

\begin{figure*}[p]
\centering
\includegraphics[width=\textwidth]{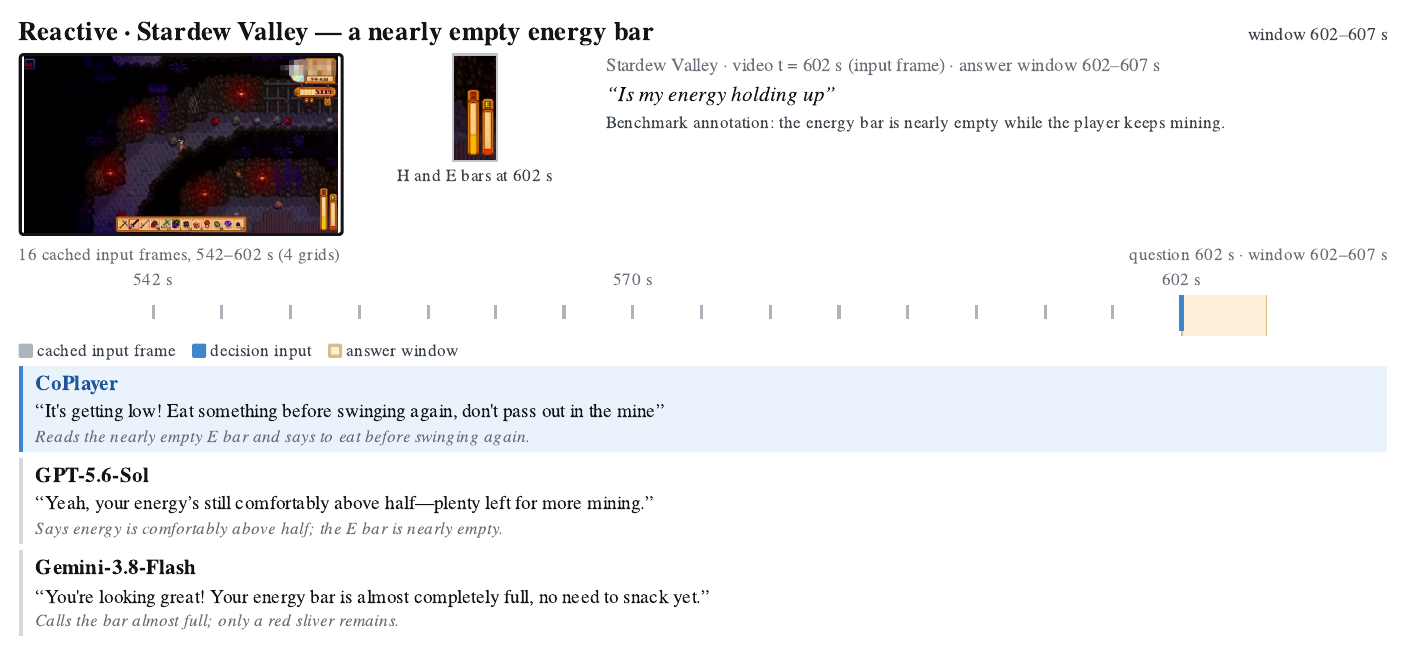}
\caption{\textbf{Stardew Valley: a nearly empty energy bar.}
The frame is the exact model input; the inset magnifies the H and E bars at the decision time.
CoPlayer reads the orange energy bar and says to eat before swinging again, while GPT-5.6-Sol calls the bar comfortably above half and Gemini-3.8-Flash calls it almost full.
The timeline shows the 16 cached model-input frames and the answer window.}
\label{fig:case-sdv}
\end{figure*}

Figures~\ref{fig:case}, \ref{fig:case-hs} and~\ref{fig:case-sdv} compare CoPlayer with two baselines on matched reactive questions; Figure~\ref{fig:case-proactive} adds two proactive interventions, and Figure~\ref{fig:cases} gives the paired reactive and proactive comparison.

\twocolumn[{%
\section{Human review of judge consistency}
\label{sec:supp-human}
\begin{minipage}{\textwidth}
\centering
\setlength{\abovecaptionskip}{4pt}
\setlength{\belowcaptionskip}{0pt}
\includegraphics[width=0.92\textwidth]{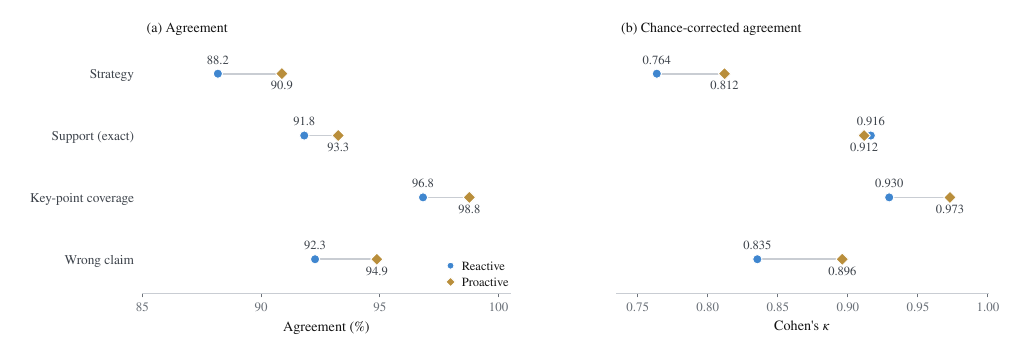}
\captionof{figure}{\textbf{Human review agrees closely with the judge in both tracks.}
(a) Agreement and (b) Cohen's $\kappa$ (support quadratic-weighted) over four criteria; blue is reactive and gold proactive; coverage is key-point agreement.}
\label{fig:human-agree}
\end{minipage}
}]

\setcounter{bottomnumber}{2}
\setlength{\textfloatsep}{8pt plus 2pt minus 2pt}
\setlength{\floatsep}{4pt plus 1pt minus 1pt}

\paragraph{Protocol and weighting.}
We sampled 40 items from the Bench split (seed 20261110), 20 reactive questions and 20 proactive events, and reviewed every model reply available on them.
The sample contains 420 unique replies (220 reactive, 200 proactive) and 597 annotator--reply rows from four raters.
Pages showed the scene and reply, translated into Chinese when needed, with no model names and in shuffled order; raters confirmed or edited the judge's prefilled strategy, support, wrong-claim and key-point labels.
The study therefore audits agreement with reviewed judge labels, not independent human ground truth; the 486 confirmed rows and 111 flagged edits (101 changing at least one field, 10 none) carry that anchoring.
Each unique reply carries total weight one, so a reply with $n$ ratings contributes $1/n$ and the six shared items do not receive four votes; support uses the fixed 0--4 scale with quadratic-weighted $\kappa$, and coverage is key-point coverage, not an F1 score.
Coverage draws on 477 reactive and 671 proactive key-point observations; the shared subset contributes 708 key-point pairs rather than 354 reply pairs.
The two tracks are analyzed separately.

\paragraph{Agreement and corrections.}
Under reply-equal weighting, agreement is 89.5\% for strategy ($\kappa=0.787$), 92.5\% for exact support (quadratic $\kappa=0.915$), 97.7\% for key-point coverage ($\kappa=0.950$) and 93.5\% for wrong claims ($\kappa=0.866$); support agreement within $\pm1$ is 96.7\%.
Per-track values in Table~\ref{tab:human_eval} differ by at most 2.7 points in agreement and 0.05 in $\kappa$; a 597-row weighting changes global values by at most 2.0 points.
Among the 101 edited rows, raters raised strategy 44 times against 7 decreases and support 47 times against 4 (mean $+1.49$ on 0--4), removed 28 flagged wrong claims against 3 additions, and added 18 key points against 1 removal.
The judge is therefore stricter than the reviewed labels on strategy, support and wrong claims, while key-point coverage is nearly saturated.

\begin{table}[b]
\centering
\caption{\textbf{Human review agrees with the judge on both tracks.}}
\label{tab:human_eval}
\footnotesize
\setlength{\tabcolsep}{7.85pt}
\begin{tabular}{l cccc}
\toprule
& \multicolumn{2}{c}{Reactive} & \multicolumn{2}{c}{Proactive} \\
\cmidrule(lr){2-3}\cmidrule(lr){4-5}
Criterion & Agr. (\%) & $\kappa$ & Agr. (\%) & $\kappa$ \\
\midrule
Strategy (yes vs.\ no) & 88.2 & 0.764 & 90.9 & 0.812 \\
Support (0--4), exact & 91.8 & 0.916 & 93.3 & 0.912 \\
Key-point coverage & 96.8 & 0.930 & 98.8 & 0.973 \\
Wrong claim & 92.3 & 0.835 & 94.9 & 0.896 \\
\bottomrule
\end{tabular}
\end{table}

\begin{figure*}[t]
\centering
\includegraphics[width=0.92\textwidth]{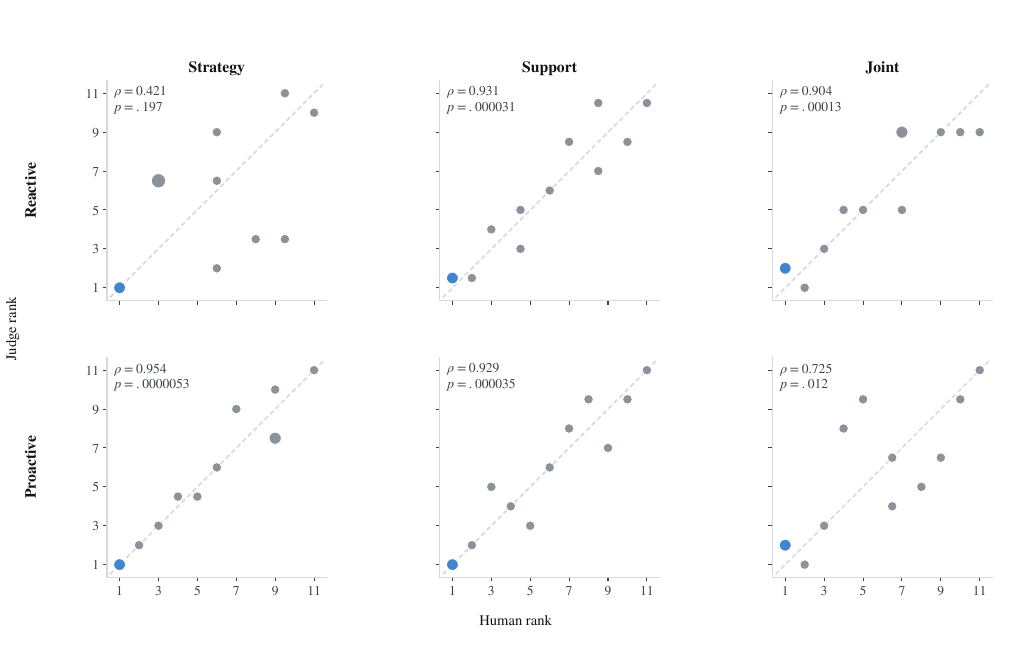}
\caption{\textbf{Human and judge model rankings agree except for reactive strategy.}
Ranks over 11 models per track for strategy, support and the joint criterion; rank 1 is best, dashed is equality, blue is CoPlayer, marker area counts shared coordinates; each panel gives $\rho$ and $p$.}
\label{fig:human-rank-grid}
\end{figure*}

\paragraph{Human--human reliability.}
Agreement among raters is measurable only on the six shared items: 59 replies (48 proactive, 11 reactive), all four raters and 354 real rater pairs.
Pairwise agreement is 96.6--100\% for strategy ($\kappa=0.932$--$1.000$), 83.1--93.2\% for exact support (fixed-pair quadratic $\kappa=0.806$--$0.949$), 99.2--100\% for coverage ($\kappa=0.981$--$1.000$) and 94.9--100\% for wrong claims ($\kappa=0.894$--$1.000$; Table~\ref{tab:human_human_pairs}).
Multi-rater ordinal Krippendorff $\alpha$ is 0.906 for support, 0.961 for strategy, 0.990 for coverage and 0.947 for wrong claims.
It is a fixed shared subset under the same prefill, not a ceiling for the full 420-reply study.
Dropping the 279 pairs both raters left untouched leaves 75 disagreement-enriched pairs with support exact agreement 45.3\% (quadratic $\kappa=0.011$) and coverage 98.0\%.
Dropping the 284 pairs neither rater changed leaves 70 pairs at 41.4\% (quadratic $\kappa=-0.042$).
These subsets show how strongly the headline values are anchored by confirmed prefills; they are descriptive, not reliability estimates.

\begin{table}[b]
\centering
\caption{\textbf{Human--human agreement on the six shared items.}
59 shared replies (118 key-point items); QWK $=$ quadratic $\kappa$.}
\label{tab:human_human_pairs}
\footnotesize
\setlength{\tabcolsep}{4.225pt}
\begin{tabular}{l ccc c}
\toprule
Pair & \makecell{Strategy\\Agr.\,(\%) / $\kappa$} & \makecell{Support\\Agr.\,(\%) / QWK} & \makecell{Coverage\\Agr.\,(\%) / $\kappa$} & \makecell{Wrong\\Agr.\,(\%) / $\kappa$} \\
\midrule
A--B & 98.3 / 0.966 & 93.2 / 0.949 & 99.2 / 0.981 & 94.9 / 0.894 \\
A--C & 98.3 / 0.966 & 84.7 / 0.832 & 99.2 / 0.981 & 96.6 / 0.929 \\
A--D & 98.3 / 0.966 & 83.1 / 0.806 & 99.2 / 0.981 & 96.6 / 0.929 \\
B--C & 96.6 / 0.932 & 89.8 / 0.844 & 100.0 / 1.000 & 98.3 / 0.965 \\
B--D & 96.6 / 0.932 & 88.1 / 0.817 & 100.0 / 1.000 & 98.3 / 0.965 \\
C--D & 100.0 / 1.000 & 91.5 / 0.892 & 100.0 / 1.000 & 100.0 / 1.000 \\
\bottomrule
\end{tabular}
\end{table}

\paragraph{Track-wise rankings and sensitivity.}
We compare judge and human means within each track over 11 models, using one judge score and the rater mean per reply; Figure~\ref{fig:human-rank-grid} plots the corresponding ranks for all criteria and tracks (ties averaged), and Figure~\ref{fig:human-agree} shows the reply-level agreement behind them.
Support rankings agree closely ($\rho=0.931$ reactive, $0.929$ proactive) and joint rankings are $\rho=0.904$ and $0.725$.
Strategy rankings correlate strongly proactively ($\rho=0.954$) but weakly reactively ($\rho=0.421$, $p=0.197$), where human and judge model means take only six and seven distinct values (tie deficits five and four).
Row-weighting or excluding shared replies leaves support and joint correlations close to primary; reactive strategy stays low under reply-equal weighting ($\rho=0.421$) and shared-excluded ($\rho=0.429$) and rises only when shared replies are row-weighted ($\rho=0.855$; Table~\ref{tab:ranking_rho}). A balanced seven-task proactive subset gives $\rho=0.833$, $0.977$ and $0.653$ for strategy, support and joint.
Pooled 12-model values from the original analysis ($\rho=0.937$, $0.895$, $0.867$ and $0.986$ for strategy, support, joint and coverage) mix track-specific policies and give shared replies four votes; we report them only as compatibility numbers, not per-track results.

\begin{table}[b]
\centering
\caption{\textbf{Track-wise judge--human rank correlations.}
Spearman $\rho$ over 11 models per track; the last two columns are sensitivity analyses. Primary correlations have $p<0.05$; the exception is reactive strategy, whose correlation is not significant ($p=0.197$).}
\label{tab:ranking_rho}
\footnotesize
\setlength{\tabcolsep}{5.48pt}
\begin{tabular}{ll cccc}
\toprule
Track & Metric & \makecell{Reply-\\equal $\rho$} & $p$ & \makecell{Row-\\weighted $\rho$} & \makecell{Excl.\\shared $\rho$} \\
\midrule
Reactive & Strategy & 0.421 & 0.197 & 0.855 & 0.429 \\
 & Support & 0.931 & $<0.001$ & 0.975 & 0.937 \\
 & Joint & 0.904 & $<0.001$ & 0.926 & 0.917 \\
\midrule
Proactive & Strategy & 0.954 & $<0.001$ & 0.890 & 0.954 \\
 & Support & 0.929 & $<0.001$ & 0.907 & 0.890 \\
 & Joint & 0.725 & 0.012 & 0.653 & 0.793 \\
\bottomrule
\end{tabular}
\end{table}

\paragraph{Scope.}
Proactive items are hit-conditioned: each event required a scored hit from at least eight systems, 13 of the 20 events contain 8--10 replies, and 14 of 20 events have model-specific hit-step spreads of 0--4 decision steps.
Reviewers therefore saw one event-level frame grid whose exact observation step can differ across replies, and the study scores emitted replies rather than whether or when a system should speak.
It does not validate missed windows, intervention timing, or event annotation content and timing.

\end{document}